\documentclass[10pt,journal,compsoc]{IEEEtran}
\ifCLASSOPTIONcompsoc
\else
\fi
\ifCLASSINFOpdf
\else
\fi
\usepackage{amsmath,amsfonts}
\usepackage{algorithmic}
\usepackage{algorithm}
\usepackage{array}
\usepackage{textcomp}
\usepackage{stfloats}
\usepackage{url}
\usepackage{verbatim}
\usepackage{graphicx}
\usepackage{float}
\usepackage{subfig}
\usepackage{cite}
\usepackage{xcolor}
\usepackage{diagbox}
\usepackage{subfloat}
\usepackage{multirow}
\usepackage{booktabs}
\usepackage{makecell}
\usepackage{hyperref}

\begin{document}

\title{A Hybrid Neural Coding Approach for Pattern Recognition with Spiking Neural Networks}

\author{Xinyi Chen{*}, Qu Yang{*}, Jibin~Wu,
	Haizhou~Li,~\IEEEmembership{Fellow,~IEEE},	        
        Kay~Chen~Tan,~\IEEEmembership{Fellow,~IEEE}

\IEEEcompsocitemizethanks{
\IEEEcompsocthanksitem This work was supported in part by the IAF, A*STAR, SOITEC, NXP and National University of Singapore under FD-fAbrICS: Joint Lab for FD-SOI Always-on Intelligent Connected Systems (Award I2001E0053), the Agency for Science, Technology and Research (A*STAR) under its AME Programmatic Funding Scheme (Project No. A18A2b0046), A*STAR under its RIE 2020 Advanced Manufacturing and Engineering Human (AME) Programmatic Grant (Grant No. A1687b0033), the Research Grants Council of the Hong Kong SAR (Grant No. PolyU11211521, PolyU15218622, and PolyU25216423), and in part by The Hong Kong Polytechnic University (Project IDs: P0039734, P0035379, P0043563, and P0046094), and in part by the National Natural Science Foundation of China (NSFC) under Grant U21A20512 and 62306259.
\IEEEcompsocthanksitem *X.~Chen and Q.~Yang contributed equally in this work. Corresponding Author: J.~Wu (jibin.wu@polyu.edu.hk)
\IEEEcompsocthanksitem X.~Chen, J.~Wu and K.~C.~Tan are with the Department of Computing, The Hong Kong Polytechnic University, Hong Kong SAR. 
\IEEEcompsocthanksitem Q.~Yang and H.~Li are with the Department of Electrical and Computer Engineering, National University of Singapore, Singapore. H. Li is also with Shenzhen Research Institute of Big Data, School of Data Science, The Chinese University of Hong Kong, Shenzhen (CUHK-Shenzhen), China. 
}
}



\IEEEtitleabstractindextext{%
\begin{abstract}
Recently, brain-inspired spiking neural networks (SNNs) have demonstrated promising capabilities in solving pattern recognition tasks. However, these SNNs are grounded on homogeneous neurons that utilize a uniform neural coding for information representation. Given that each neural coding scheme possesses its own merits and drawbacks, these SNNs encounter challenges in achieving optimal performance such as accuracy, response time, efficiency, and robustness, all of which are crucial for practical applications. In this study, we argue that SNN architectures should be holistically designed to incorporate heterogeneous coding schemes. As an initial exploration in this direction, we propose a hybrid neural coding and learning framework, which encompasses a neural coding zoo with diverse neural coding schemes discovered in neuroscience. Additionally, it incorporates a flexible neural coding assignment strategy to accommodate task-specific requirements, along with novel layer-wise learning methods to effectively implement hybrid coding SNNs. 
We demonstrate the superiority of the proposed framework on image classification and sound localization tasks. Specifically, the proposed hybrid coding SNNs achieve comparable accuracy to state-of-the-art SNNs, while exhibiting significantly reduced inference latency and energy consumption, as well as high noise robustness. This study yields valuable insights into hybrid neural coding designs, paving the way for developing high-performance neuromorphic systems.
\end{abstract}

}

\maketitle

\section{Introduction}

Human brains are the most sophisticated yet energy-efficient computational systems, wherein biological neurons use electrical impulses or `spikes' to transmit information to each other. Spiking neural networks (SNNs), with rich neuronal dynamics and sparse spiking activities, are designed to mimic the remarkable information processing mechanism of biological neural systems \textit{in silico}. The higher degree of biological details encompassed endows SNNs with a greater potential to reach the unprecedented performance of their biological counterparts. 

{Recent advancements in SNN training algorithm development \cite{wu2018spatio, 8891809, wu2021tandem} and model architecture design \cite{zheng2021going, fang2021deep} have enabled a wide range of applications, including image classification \cite{qin2023attention, wang2023adaptive}, speech processing \cite{wu2018spiking, wu2020deep, wu2021progressive}, and robotic control \cite{dewolf2021spiking, BDETT}, etc. However, it remains elusive how to encode sensory signals into spike trains and process them within the SNN to achieve optimal task performance. The state-of-the-art (SOTA) SNNs used to solve pattern recognition tasks are primarily bootstrapped from homogeneous neurons, that share a uniform neural coding scheme. It further relies on extended learning with large-scale datasets to reach high classification accuracies. This homogeneous coding approach encounters challenges in effectively achieving multiple objectives that a system is expected to fulfill in practice, encompassing accuracy, efficiency, response time, and robustness. For instance, consider a robotic system integrating diverse sensory inputs like vision, touch, and sound. The signals collected at different spatial and temporal resolutions demand disparate treatment.
}

The findings in biological neural systems prompt us to look into diversified neural coding schemes in neural computation. Over the last few decades, growing biological evidence indicates that perception and cognition are supported by diversified neural coding schemes, which are believed to play an essential role in signal processing, learning, and memory in neural systems \cite{gjorgjieva2016computational}. The earliest research on neural coding can be traced back to 1926 when Adrian and Zotterman first revealed that sensory nerves innervating the muscle discharge spikes at a rate proportional to the weights hung from the muscle \cite{rate-coding}. Since then, substantial evidence has been accumulated from different neural systems and led to the conclusion that the firing rate of neurons is the key information carrier. However, the rate coding failed to account for many important features of neural computation. To address this limitation, temporal coding schemes have been proposed as complementary approaches to explain the limitations of rate coding. For example, in the human visual system, object recognition can be accomplished within 150 ms since the stimulus onset \cite{visual}, during which signals need to travel across multiple interconnected stages. The rank order coding has been proposed to explain this rapid information processing. It suggests the sensory information is represented in the chronological order of spikes, and more important features will be conveyed earlier \cite{thorpe2001spike}. Additionally, the spiking activities are found to be phase-correlated with the ongoing neural oscillations during memory formulation in the hippocampus \cite{hippo}. Moreover, as observed in the thalamus, hippocampus, and auditory systems, neurons are capable of firing a burst of spikes with a small inter-spike interval. This rapid information transmission mechanism has led to the development of burst coding \cite{burst}. 

\begin{figure*}[htb]
\centering
\includegraphics[scale=1.1, trim=0 0 0 0,clip]{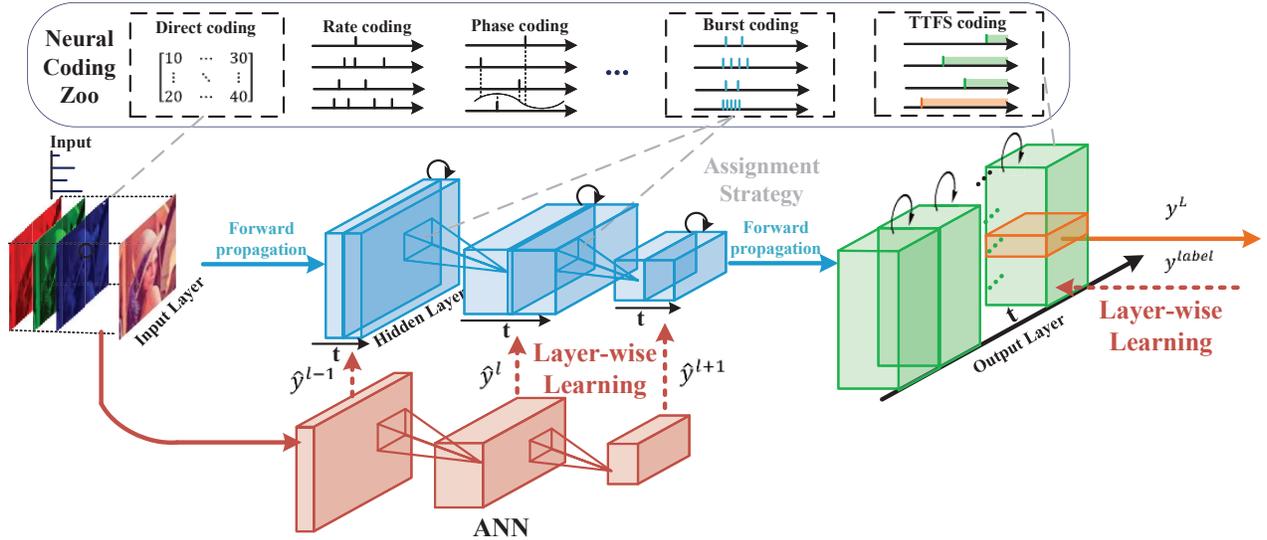}
\vspace{-2mm}
\caption{{Illustration of the proposed hybrid neural coding and learning framework in solving the image classification task. The proposed framework consists of three components: Neural Coding Zoo (upper box), Assignment Strategy (gray dotted line), and Layer-wise Learning Method (red dotted line). First of all, the neural coding zoo is a comprehensive collection
of the most representative neural coding schemes discovered in neuroscience. According to the specific task requirements of image classification, the assignment strategy has been further designed to assign heterogeneous neural coding schemes from the neural coding zoo to different network layers: direct coding for high-fidelity input representation, burst coding for reliable and rapid hidden feature representation, and TTFS coding to ensure fast and efficient output decision-making. Finally, the layer-wise learning methods are introduced to independently train the hidden and output layers, so as to achieve the desired coding schemes. Here, $\boldsymbol{y}^L$ represents the outputs of the hybrid coding SNN, $\boldsymbol{y}^{label}$ denotes the one-hot labels, $\hat{\boldsymbol{y}}^l$ signifies the activation values of neurons at hidden layer $l$ in the teacher ANN.}}
\label{fig: coding-scheme}
\end{figure*}

{The large body of research on neural coding has provided a wealth of inspiration for the SNN design. Nevertheless, most of the existing SNNs are only grounded on a single type of these neural coding schemes, overlooking the critical interplay between them. Consequently, this design methodology typically leads to suboptimal performance for the targeted task.
For instance, direct coding uses the analog value for the input or output layer, maintaining a high resolution of information but is the least biologically plausible and energy-efficient scheme. The majority of research on ANN-to-SNN conversion employs rate coding to approximate the ReLU activation function output with the firing rate of spiking neurons \cite{cao2015spiking, QCFS, Calibration,ijcai2021-321,deng2021optimal}. However, the high accuracy of the converted SNNs often comes at the cost of long inference time. As a more reliable coding scheme, burst coding encodes information into the inter-spike interval (ISI) and allows a burst of spikes to be transmitted within a short time interval for robust information transmission \cite{2019fast}. Despite the superior performance, implementing burst coding requires a high computational cost.
Some recent studies are looking at more efficient coding schemes. Phase coding, for example, embeds information into the spike-phase-based neural oscillations, which remarkably reduces the spike density \cite{phase}. However, primitive phase coding struggles to represent a highly dynamic information stream. The time-to-first-spike (TTFS) coding, on the other hand, encodes information via the first spike time \cite{T2FSNN,zhang2021rectified, TDSNN}. However, these TTFS-based SNNs face gradient instability and dead neuron problems during training \cite{zhang2021rectified}, which restricts its scalability.}

{In this paper, we argue that SNNs should be holistically designed by incorporating various neural coding schemes to achieve optimal performance.} To this end, we propose a hybrid neural coding and learning framework. As illustrated in Fig. \ref{fig: coding-scheme}, the proposed framework incorporates a neural coding zoo formed by a collection of bio-inspired neural coding schemes. According to the assignment strategy and task requirements, these coding schemes are allocated to each network layer, thereby providing the desired information representation basis. Moreover, we introduce a layer-wise learning method to ensure the desired neural representation can be effectively achieved at each network layer. {Consequently, the hybrid coding SNNs could take advantage of diversified coding schemes to achieve optimal task performance.} We verify the effectiveness of the proposed framework on both image and audio perception tasks, and the results demonstrate the desired attributes of high accuracy and robustness, as well as low latency and energy consumption. The main contributions of this paper are summarized as follows:

\begin{itemize}

\item {We present a novel hybrid neural coding and learning framework, which takes a holistic approach to design the neural coding schemes of an SNN so as to meet the specific task requirements and environment conditions.}
\item {We propose a layer-wise learning method that can efficiently achieve the desired neural coding schemes with high levels of efficacy.}
\item {We showcase the superiority and versatility of the proposed framework in performing image classification and sound localization tasks.
}
\item Our research yields valuable insights into diverse hybrid neural coding designs, which serve as practical guidelines for solving real-world pattern recognition tasks with hybrid coding SNNs. 
\end{itemize}

The remainder of this paper is organized as follows. In Section 2, we present the proposed hybrid neural coding and learning framework in detail. In Sections 3 and 4, we evaluate the performance of the proposed framework on image classification and sound localization tasks, respectively. Additionally, we conduct a series of ablation studies in Section 4 to compare different hybrid coding designs. Finally, we conclude our findings in Section 5.

\section{A Hybrid Neural Coding and Learning Framework for Spiking Neural Networks}
\label{sec:method}
In this section, we present our proposed hybrid neural coding and learning framework that provides a holistic way to design hybrid coding SNNs. As illustrated in Fig. \ref{fig: coding-scheme}, our framework consists of three key components: neural coding zoo, hybrid coding assignment strategy, and layer-wise learning method. In the following, we begin with an overview of SNNs and their coding schemes. Building upon this, we then describe the collection of brain-inspired neural coding schemes that jointly form our neural coding zoo. We also elaborate on the strategy employed to assign these coding schemes to different layers of the network. Furthermore, we take image classification and sound localization tasks to illustrate how a hybrid coding assignment strategy could be designed in practice to fulfill distinct task objectives. Finally, we present the proposed layer-wise learning method.

\subsection{Neural Coding Zoo}
{Unlike traditional artificial neurons, spiking neurons are designed to reproduce the complex neuronal dynamics of biological neurons. Notably, spiking neurons use spike trains as information carriers to transmit information to each other. The method of how information is represented by a series of spike trains, a concept referred to as a neural coding scheme, significantly influences its information processing capability. In general, the frequency, precise spike timing, and synchrony patterns within the spike trains could all be utilized to carry information. An appropriate assignment of these neural coding schemes to an SNN could significantly enhance its information processing efficiency, efficacy, and robustness.}

\begin{figure}[htb]
    \centering
    \includegraphics[scale=0.52,trim=70 30 5 18,clip]{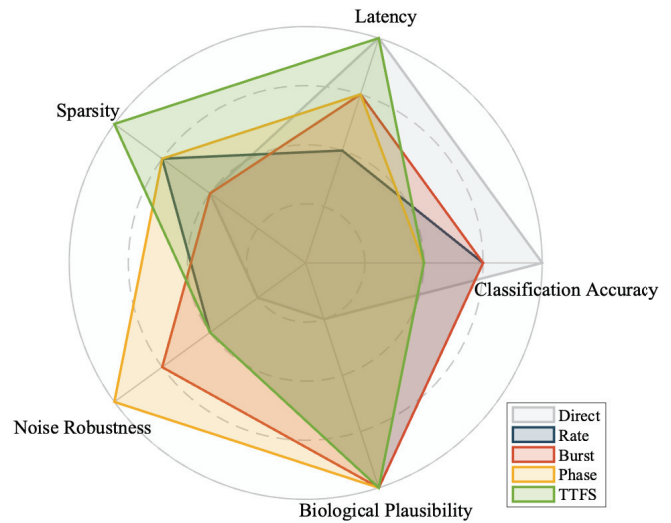}
    \caption{Comparison of computational performance and biological plausibility among diverse candidate neural coding schemes.}
    \label{fig:coding_compare}
\end{figure}

Inspired by the rich biological neural representations uncovered in neuroscience, a variety of neural coding schemes, each with its unique strengths and limitations, have been explored for SNNs. In our proposed framework, the neural coding zoo is a comprehensive collection of the most representative neural coding schemes, which provides a wealth of candidate options for designing hybrid coding SNNs. It includes direct coding, rate coding, phase coding, burst coding, and TTFS coding, facilitating the construction of versatile SNNs that can meet diversified task requirements and adapt to various environmental conditions. In the following, we provide a detailed summary of each candidate neural coding scheme, with their merits and weaknesses depicted in Fig. \ref{fig:coding_compare}. It is worth noting that our neural coding zoo can effortlessly integrate any newly discovered or developed neural coding schemes. 

\begin{itemize}
\item \textbf{Direct coding} has been widely used in input and output layers of SNNs, which directly employs analog values to encode input stimuli or to decode outputs \cite{wu2021tandem, wu2021progressive}. Given its analog nature, it is the least biologically plausible and energy-efficient coding scheme, as it abandons the binary spike-based information representation. Nevertheless, direct coding could avoid sampling errors and delays inherent in other coding schemes, thus ensuring a high fidelity in information representation. 
\item \textbf{Rate coding} employs the mean firing rate within a given time window to signify the signal value. Due to the similarity between ReLU activation values and firing rates of spiking neurons, rate coding has been widely adopted in ANN-to-SNN conversion methods \cite{cao2015spiking, QCFS, Calibration,ijcai2021-321,deng2021optimal} to construct SNNs that can well approximate the feature representation of ANNs. Despite its biological plausibility and superior information representation capability, rate coding is often criticized for high latency, poor energy efficiency, and ineffectiveness in representing highly dynamic signals. 
\item \textbf{Phase coding} represents information through the phase of spiking activities relative to underlying neural oscillations \cite{neural_pop}. It is proven to be more efficient in information transmission than rate coding and offers higher noise robustness than TTFS coding \cite{spike-phase}. However, the representation capacity of phase coding is constrained by the frequency of underlying neural oscillations, which leads to reduced performance when tasked with representing rapidly changing stimuli \cite{KIM2018373, phase}.

\item \textbf{Burst coding} makes effective use of the ISI to represent information \cite{IZH_burst} and is highly efficient in information transmission \cite{neuralcoding}. However, keeping track of the ISI introduces a high computational complexity. To address this issue, recent works propose using a neuron model that can emit multiple spikes within a single timestep to represent graded ISIs \cite{2019fast,li2022efficient}, resulting in a simple yet robust neural coding scheme. 

\item \textbf{TTFS coding} encodes information based on the timing of the first spike, where earlier spikes represent more significant information. This coding scheme leads to the favorable properties of fast inference, low spike density, and low energy consumption \cite{sparse_temporal,Oh_2022}. However, TTFS coding suffers from the notorious dead neuron problem, which can affect the training stability. Furthermore, spike timing is highly sensitive to neuronal noise, a common occurrence in neural systems and analog computing substrates. 

\end{itemize}

\subsection{Hybrid Coding Assignment Strategy}
\label{sec:coding}
As discussed earlier, each homogeneous neural coding scheme possesses its unique merits and weaknesses, and if applied alone, it will significantly impact the overall performance of SNNs. {Therefore, our objective is to design hybrid coding SNNs that can leverage the benefits of these homogeneous neural coding schemes to achieve optimal task performance.}
This can be achieved by selecting appropriate neural coding schemes from our neural coding zoo and assigning them to different parts of the network according to varying task requirements and environmental conditions. 
However, this will lead to a complex combinatory optimization problem that is intractable for large-scale networks. To reduce the design effort and accelerate training, we organize the network layers into three blocks, each of which adopts a homogeneous neural coding scheme throughout its entirety. Specifically, we focus on three network blocks in this work: Input Layer, Hidden Layers, and Output Layer. In the following, we describe the critical factors when assigning neural coding schemes to these three network blocks.

\begin{itemize}
\item \textbf{Input Layer} plays a significant role as the interface between an SNN and its external environment. It is crucial for this layer to provide an efficient, real-time, and high-fidelity representation of the input stimuli. To achieve these objectives, the input coding scheme must be tailored to the nature of the input signal and the prevailing environmental conditions, ensuring a fast response while minimizing information loss during the sampling and encoding process.
\item \textbf{Hidden Layers} play a pivotal role in extracting rich and hierarchical feature representations from the input stimuli. Consequently, it is imperative that the chosen coding scheme not only facilitate efficient feature extraction but also balance between computational cost and robustness to neuronal noise.
\item \textbf{Output Layer} functions as the output interface for decision generation, thereby underscoring the need for a neural coding scheme that can simultaneously facilitate accurate and fast decision-making.
\end{itemize}

This assignment strategy serves as general guidelines on how to construct an optimal hybrid coding SNN that can meet task-specific objectives and environmental conditions. 

\subsection{Hybrid Coding Design for Image Classification}
{In this section, grounded on the neural coding zoo and assignment strategy introduced earlier, we will illustrate how a hybrid coding SNN can be practically constructed to tackle the image classification task.}

\begin{figure*}[htb]
\centering
\includegraphics[scale=0.74, trim=0 0 0 0,clip]{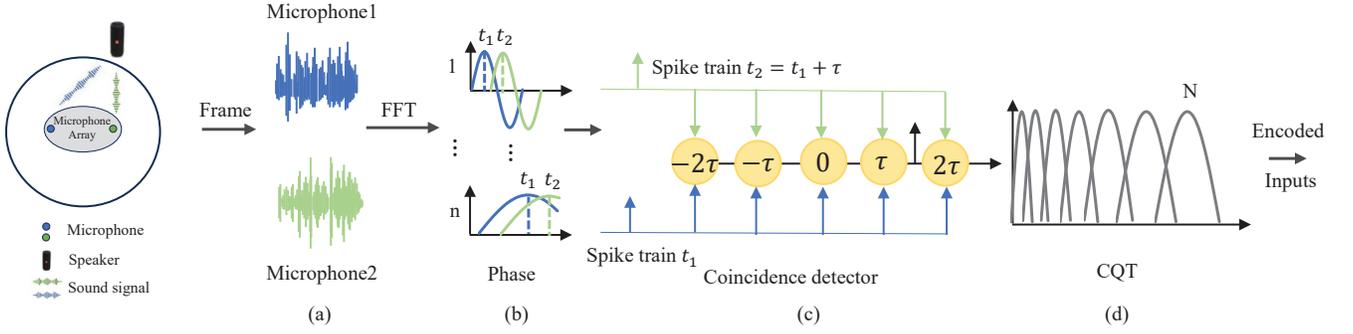}
\vspace{-2mm}
\caption{Illustration of the phase encoding front-end used for the sound localization task, where two microphones are selected for demonstration. \textbf{(a)} The raw audio is decomposed into $n$ single-tone sinusoidal signals by fast Fourier transform (FFT). \textbf{(b)} The first peak time of each sinusoidal signal is encoded into a precise-timing spike, which represents the arrival time of each sound. \textbf{(c)} $N_{\tau}$ coincidence detection neurons are employed to detect the phase difference between two microphones. If a coincidence detection neuron detects its predefined phase difference, it emits a spike. \textbf{(d)} To reduce the computational cost of post-processing, the encoded spike trains of the $n$ single tones are further grouped into $N_c$ channels via constant Q transform (CQT).}
\label{fig: prepocess}
\end{figure*}

\subsubsection{Direct Coding for High-Fidelity Input Representation}
In image classification tasks, a high-fidelity representation of input images is typically required. However, the use of rate coding may require a large sampling time window and high firing rates, leading to slow and inefficient runtime. On the other hand, temporal coding schemes (e.g., TTFS coding) may encounter issues like dead neurons and gradient instability, adversely affecting the ease of training \cite{zhang2021rectified}. To address these challenges, we adopt a direct coding scheme that treats the intensity value of image pixels as the input current and directly injects it into the neurons in the first hidden layer \cite{wu2021tandem, wu2021progressive}. This approach bypasses the analog-to-spike conversion process, eliminating the information loss during rate-based sampling as well as the quantization errors during temporal encoding. 
\subsubsection{Burst Coding for Reliable and Rapid Hidden Feature Representation}
 In contrast to rate coding, which restricts neurons to emit at most one spike at a time, bursting neurons can generate up to $\Gamma$ spikes. This characteristic facilitates faster transmission of important information with greater robustness. Therefore, we employ burst coding to hidden layers to ensure rapid and efficient transfer of spike-based feature representations to subsequent layers. {Specifically,  $\Gamma$ is set to 5 considering that neurons rarely emit a spike burst with more than five spikes in a biological discovery \cite{buzsaki2012neurons} (see more detailed analysis on the selection of $\Gamma$ in Supplementary Materials)}. We utilize the bursting neuron model introduced in \cite{li2022efficient}, in which the membrane potential is charged as:
\begin{equation}\label{IF_mem}
\boldsymbol{v}^l(t)=\boldsymbol{v}^l(t-1) + {I^l(t)},
\end{equation}
where ${I^l(t)}$ denotes the input current induced from the input spikes, which is defined as ${I^l(t)}=\mathcal{W}^l\boldsymbol{s}^{l-1}(t)$. $\boldsymbol{s}^{l-1}(t)$ is the input spikes from layer $l-1$, $\mathcal{W}^l$ represents the weight matrix between layer $l$ and $l-1$. Once the membrane potential $\boldsymbol{v}^l$ exceeds the firing threshold $V^l_{th}$, an output spike burst $\boldsymbol{s}^l$ will be generated, and the membrane potential $\boldsymbol{v}^l$ will be reset using the reset-by-subtraction scheme \cite{rueckauer2017conversion} as:
\begin{equation}
\begin{aligned}
\boldsymbol{s}^l(t)&= \operatorname{min} \left(
\operatorname{max}\left(\lfloor \frac{\boldsymbol{v}^l(t)}{V^l_{th}} \rfloor, 0\right), \Gamma\right),\\
\boldsymbol{v}^l(t)&=\boldsymbol{v}^l(t) - \boldsymbol{s}^l(t) \cdot V^l_{th},
\label{eq: burst}
\end{aligned}
\end{equation}
where the $\boldsymbol{s}^l(t)$ signifies the bursting spike count in the timestep $t$, whose value is taken from range $[0, \Gamma]$, 

\subsubsection{TTFS Coding for Fast and Efficient Output Decision-making}
Subsequent to the image feature extraction process in the hidden layers, the output layer is responsible for decision-making.
Ideally, the classification decision should be made as soon as sufficient evidence has been accumulated at the output layer $L$. To achieve this, we adopt TTFS coding that reaches the final decision once the first output spike has been detected, without waiting for all simulation timesteps to finish. This design thus leads to a significant reduction in inference time {while effectively circumventing the gradient instability issue commonly associated with temporal coding, as it is confined to a single-layer implementation.} Specifically, we adopt the double exponential leaky integrate-and-fire (LIF) spiking neuron model. It offers richer temporal dynamics than the bursting neuron model used in hidden layers and is defined as \cite{gutig2006tempotron} :
\begin{equation}\label{tempotron}
\begin{aligned}
\boldsymbol{v}^L(t)=&\mathcal{W}^{L} K^L(t),\\
\boldsymbol{s}^L(t)=&H(\boldsymbol{v}^L(t)-V_{th}^L),
\end{aligned}
\end{equation}
where $H(\cdot)$ denotes the Heaviside step function. Once the membrane potential $\boldsymbol{v}_i^{L}$ crosses the firing threshold $V_{th}^{L}$, the neuron emits a spike. ${K}^{L}(t)$ is the post-synaptic potential (PSP) kernel induced from input spikes as \cite{gutig2006tempotron}: 
\begin{equation}\label{psp}
\begin{aligned}
\left\{\begin{array}{lr}
{K}^L(t)=K_{0} \cdot (\boldsymbol{m}^L(t)-{I}^L(t)),&\\
\boldsymbol{m}^L(t)=e^ {-\frac{1}{\tau_{m}}} \cdot (\boldsymbol{m}^L(t-1) + \boldsymbol{s}^{L-1}(t)),&\\
{I}^L(t)=e^ {-\frac{1}{\tau_{s}}} \cdot ({I}^L(t-1) + \boldsymbol{s}^{L-1}(t)),&\\
\end{array}\right.
\end{aligned}
\end{equation}
where $K_0$ is a normalization constant that aligns the maximum value of ${K}^L$ with $V_{th}^{L}$, $\tau_m$ and $\tau_s$ are decay constants of two synaptic components $\boldsymbol{m}^L$ and ${I}^L$, respectively.

The classification result of the input images, namely $\boldsymbol{y}^{L}$, is determined by the output neuron $i$ that produces the first spike. We note the trigger time of this first spike as $t_f$. However, in some edge cases, a specific set of criteria is used to make the final decision. If multiple neurons fire simultaneously, the decision is made as follows:
\begin{equation}\label{TTFS}
\boldsymbol{y}^{L}=\max_{i}(\boldsymbol{v}_i^{L}(t_f)).
\end{equation}

Furthermore, in cases where no neuron fires until the end, we treat the last timestep as the decision time and determine the classification result according to Eq. (\ref{TTFS}).

\subsection{Hybrid Coding Design for Sound Localization}
{In this section, we further employ the proposed hybrid neural coding and learning framework to solve the sound localization task, demonstrating its generalizability across different signal modalities and task requirements.}

\subsubsection{Phase Coding for High-fidelity Input Representation}
Mammals possess a remarkable capability to figure out the origin of the sound, as they can discern the time difference of arrival (TDOA) between two ears \cite{jeffress1948place}. Specifically, sound from the same source propagates and reaches two ears with different time delays and energy attenuation, which serve as key cues for sound localization. To detect these discrepancies, each cochlea first employs phase coding to encode the disparate sound into phase-locked spike trains. Subsequently, the detection neurons in the medial superior olive (MSO) can discern the phase differences between these spike trains, thus enabling sound localization \cite{goodman2010spike}. 

Drawing inspiration from the temporal characteristics of sound signals and the biological information representation mechanism, phase coding emerges as an ideal choice for extracting and encoding localization cues. Therefore, we adopt the multi-tone phase coding (MTPC) \cite{Multitone} as the input coding scheme within our framework to ensure a high-fidelity representation of the input. As depicted in Fig. \ref{fig: prepocess}, the received audio is first decomposed into several single-tone sinusoidal signals. Then, a spike is generated at the first peak time of each sinusoidal signal, encoding the arrival time of the sound. The coincidence detection neurons then detect the phase difference between two microphones, generating an output spike if a specific time delay is observed. Finally, the output spikes generated from sub-bands are aggregated using constant Q transform (CQT) to compress the input dimensions.

\subsubsection{Burst Coding for Robust Hidden Feature Representation}
In audio signal processing, rapid information transmission and robust information representation are two desirable attributes when selecting a coding scheme for hidden layers. Low latency can significantly enhance the overall robustness of the system by allowing it to respond swiftly to changes in the input signals. Concurrently, robust information representation can effectively overcome the ubiquitous neuronal noises. To this end, we select burst coding for hidden layers, as it demonstrates superior performance in these two aspects, as shown in Fig. \ref{fig:coding_compare}.

\subsubsection{TTFS Coding for Fast and Efficient Output Decision-Making}
High accuracy, swift response to moving sound sources, and superior energy efficiency are critical attributes for any auditory interface. These factors motivated our choice of TTFS coding for the output layer of the localization network. In Section \ref{sec:effective_training}, we will illustrate how TTFS coding effectively reduces inference time without compromising localization accuracy. Furthermore, in Section \ref{sec:ablation}, we will compare TTFS coding with other coding schemes to further validate its effectiveness in the context of sound localization.

\subsection{A Layer-wise Learning Method for Hybrid Coding}
\label{sec:learning}
{Despite the flexibility that hybrid coding SNNs offer in assigning different coding schemes across various network stages, it presents its own set of challenges. Particularly, the use of different neural coding schemes renders the conventional end-to-end training approach incompatible. To resolve this challenge, we propose a novel layer-wise training method that is tailored to accomplish the designed hybrid coding SNN. In particular, the hidden and output layers are trained independently in a more manageable way.}

\subsubsection{Hidden Layer} 
\label{sec:hidden}
For all hidden layers, a layer-wise Teacher-Student learning approach is adopted. Specifically, our method leverages a pre-trained ANN as a teacher, to supervise the training of a student SNN. Considering the intermediate feature representations of the ANN as the targets, the hidden layers of our hybrid coding SNNs are trained to generate equivalent spike-count-based feature representations. For this purpose, the mean square error (MSE) between ANN and SNN features is utilized as the local loss function for each hidden layer, which is defined as:
\begin{equation}
\mathcal{L}^l\left(\hat{\boldsymbol{y}}^l, {c}^l(T)\right) = \left|\left|\hat{\boldsymbol{y}}^l - \boldsymbol{r}^l\cdot\frac{\boldsymbol{c}^l(T)}{T} \right|\right|_2^2,
\label{eq: LTL_loss}
\end{equation}
where $\hat{\boldsymbol{y}}^l$ denotes the activation of neurons in ANN layer $l$. $T$ is the time window length and $\boldsymbol{c}^l(T) = \sum^T_{t=1}\boldsymbol{s}^{l}(t)$ is the total spike count. {$\boldsymbol{r}^l$ is a layer-wise scaling factor whose value is fixed to 1. This ensures the firing rate of spiking neurons, denoted as $\boldsymbol{c}^l(T)/T$, can effectively align with the activation value of ANN neurons (see more detailed analysis on the selection of $\boldsymbol{r}^l$ in Supplementary Materials)}. 

To eliminate the mismatch between training and testing, we train each hidden layer by taking into consideration the temporal dynamics of input spike trains rather than only considering the mean firing rate as in \cite{Calibration}. Specifically, the backpropagation through time (BPTT) algorithm is employed to perform temporal credit assignments at each layer. The weight update can be derived as follows:

\begin{equation}
\begin{aligned}
\Delta \mathcal{W}^{l} \propto \frac{\partial \mathcal{L}^l}{\partial \mathcal{W}^l} = \sum_{t=1}^{T}\frac{\partial \mathcal{L}^l}{\partial \boldsymbol{v}^{l}(t)}\frac{\partial \boldsymbol{v}^{l}(t)}{\partial \mathcal{W}^l} = \sum_{t=1}^{T}\frac{\partial \mathcal{L}^l}{\partial \boldsymbol{v}^{l}(t)}\boldsymbol{s}^{l-1}(t).
\label{eq: LTL_dev}
\end{aligned}
\end{equation}

For the ease of expression, we let $\delta^l(t)= \frac{\partial \mathcal{L}^l}{\partial \boldsymbol{s}^l(t)}$ and yield 
\begin{equation}\frac{\partial \mathcal{L}^l}{\partial \boldsymbol{v}^{l}(t)} =
\begin{cases}
\delta^l(t+1) \frac{\partial \boldsymbol{s}^l(t+1)}{\partial \boldsymbol{v}^l(t+1)} + \delta^l(t) \frac{\partial \boldsymbol{s}^l(t)}{\partial \boldsymbol{v}^l(t)}, \hspace{4.0mm} \text{if} \hspace{1.5mm} t < T,\\ 
\delta^l(T) \frac{\partial \boldsymbol{s}^l(T)}{\partial \boldsymbol{v}^l(T)}, \hspace{3.2cm} \text{if} \hspace{1.5mm} t = T,\\
\end{cases}
\label{eq: dev_mem}
\end{equation}
where
\begin{equation}
    \delta^l(t)=
    \begin{cases}
    -V^l_{th} \delta^l(t+1)\frac{{\partial \boldsymbol{s}^l(t+1)}}{\partial \boldsymbol{v}^l(t+1)}+ \delta^l(T),  \hspace{3.5mm} \text{if} \hspace{1.5mm} t < T,\\ 
    -\frac{2}{T} \left(\hat{\boldsymbol{y}}^l - \boldsymbol{r}^l(\frac{1}{T} \Sigma_{t=1}^{T}\boldsymbol{s}^l(t)) \right),  \hspace{9.5mm} \text{if} \hspace{1.5mm} t = T.\\
\end{cases}
\label{eq: delta}
\end{equation}

To address the discontinuity in burst neurons as Eq. (\ref{eq: burst}), we adopt the straight-through estimator (i.e., $d\lfloor \boldsymbol{x} \rfloor/d\boldsymbol{x}=1$) \cite{bengio2013estimating} and result in $\frac{\partial \boldsymbol{s}^{l}(t)}{\partial \boldsymbol{v}^{l}(t)} = 1$. This local learning methodology, henceforth referred to as local tandem learning (LTL), broadens the scope of the original rate-coding-focused LTL rule \cite{LTL} to accommodate additional neural coding schemes.

\subsubsection{Output Layer}
\label{sec:output}
\begin{figure*}[t]
\centering
\subfloat[VGG-16 on CIFAR-10]{
	\includegraphics[scale=0.42,trim=0 20 0 14,clip]{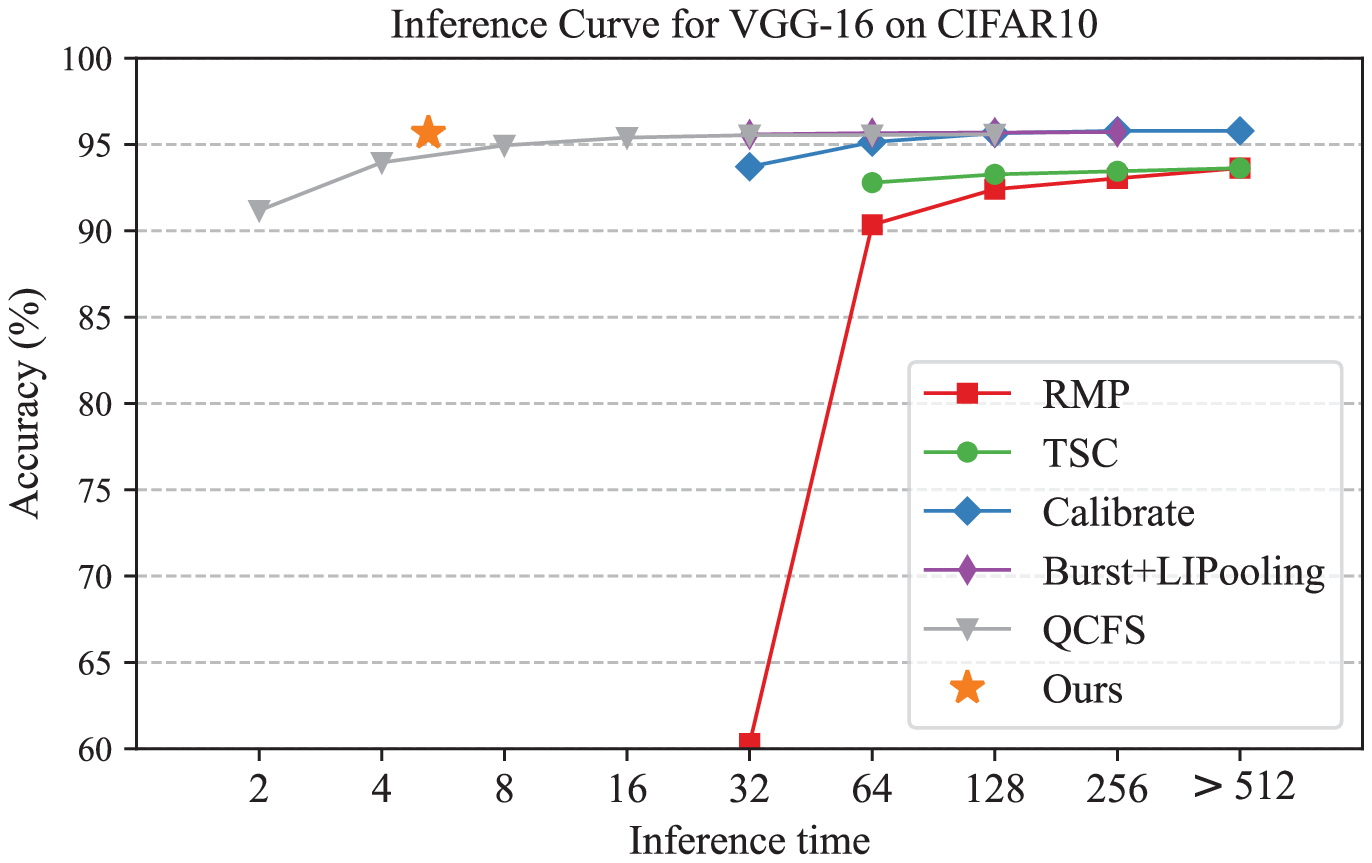}}
\subfloat[VGG-16 on CIFAR-100]{
	\includegraphics[scale=0.42,trim=0 20 0 14,clip]{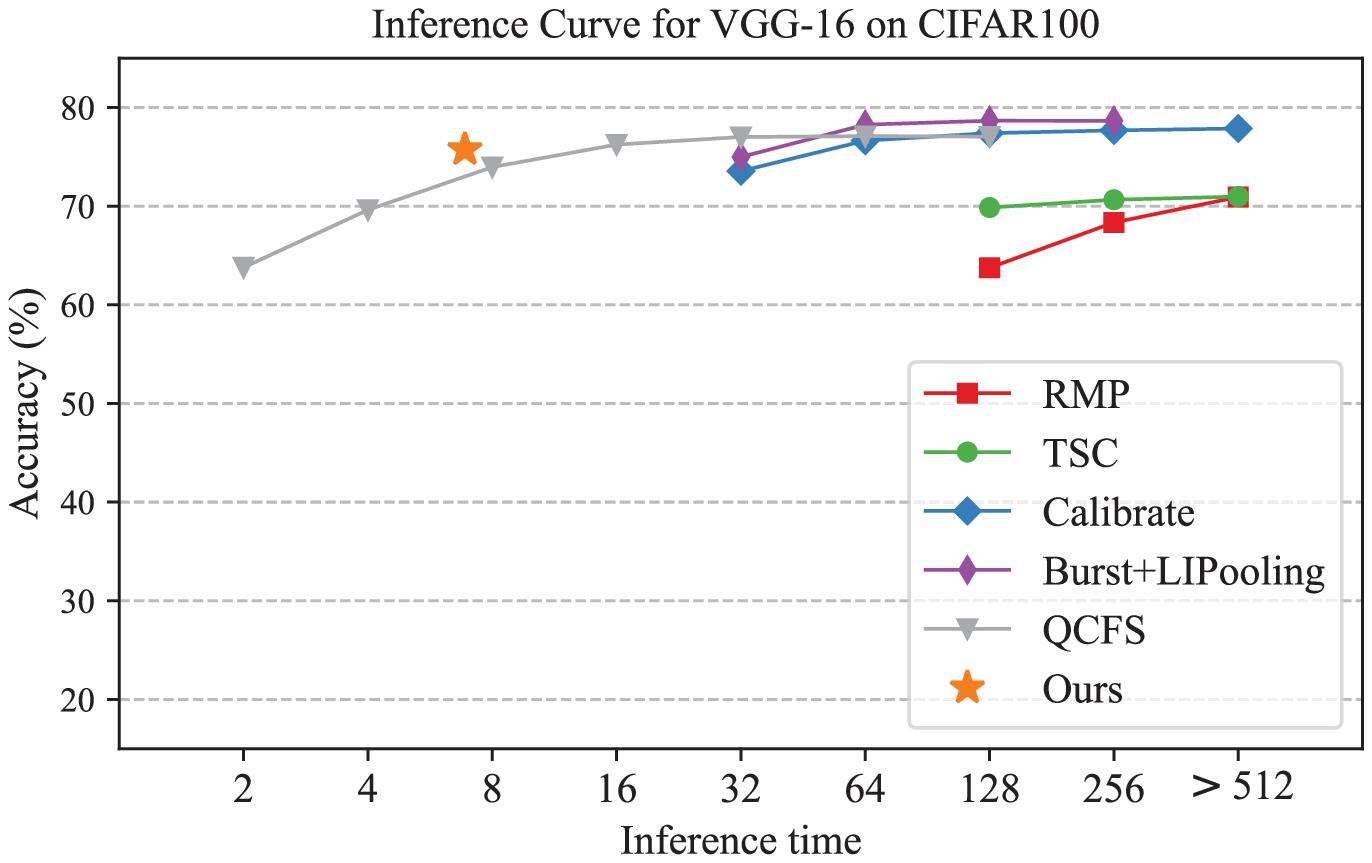}} 
 \subfloat[VGG-16 on Tiny-ImageNet]{
	\includegraphics[scale=0.42,trim=0 20 0 14,clip]{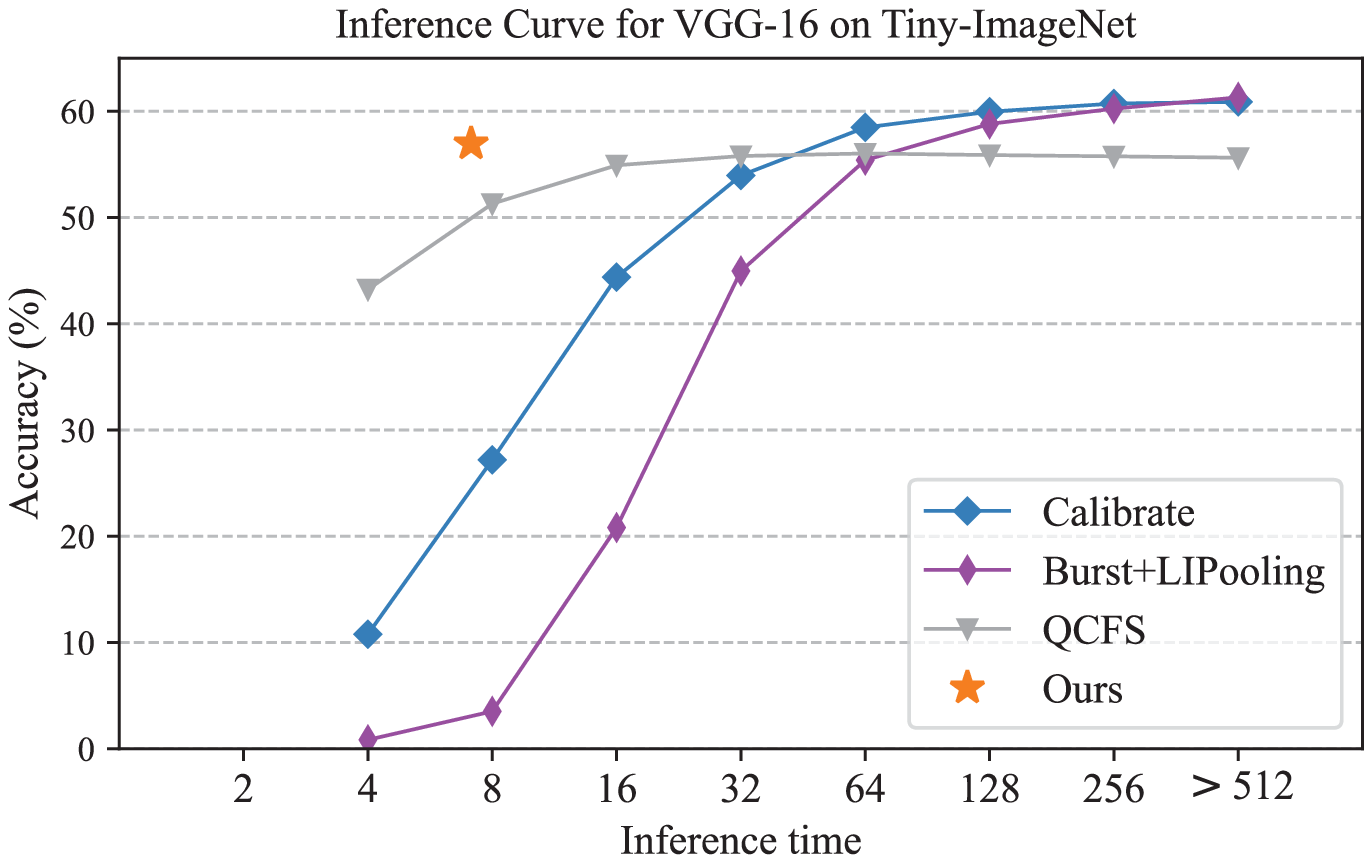}} \vspace{-3mm}\\
\subfloat[ResNet-20 on CIFAR-10]{
	\includegraphics[scale=0.42,trim=0 5 0 14,clip]{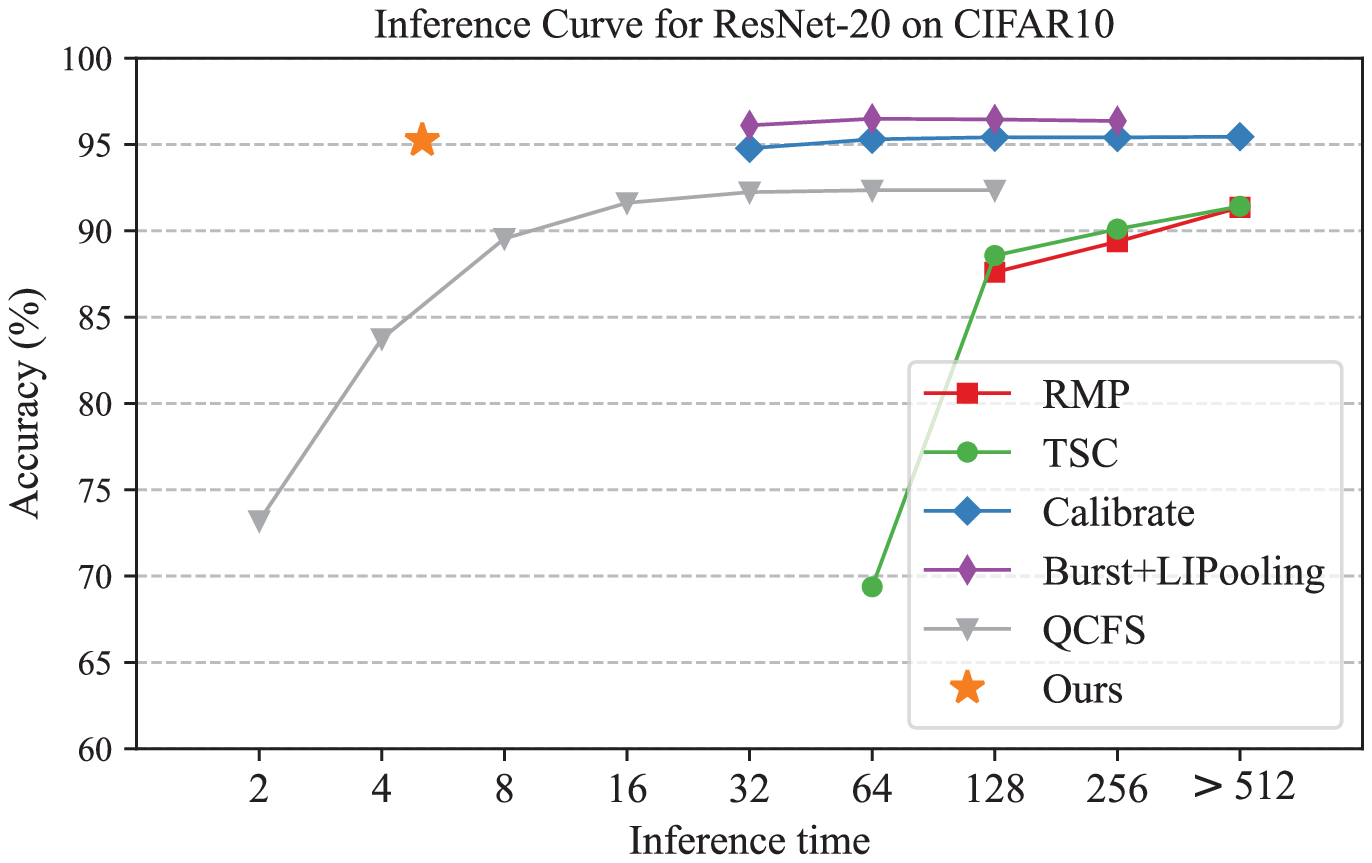}} 
\subfloat[ResNet-20 on CIFAR-100]{
	\includegraphics[scale=0.42,trim=0 5 0 14,clip]{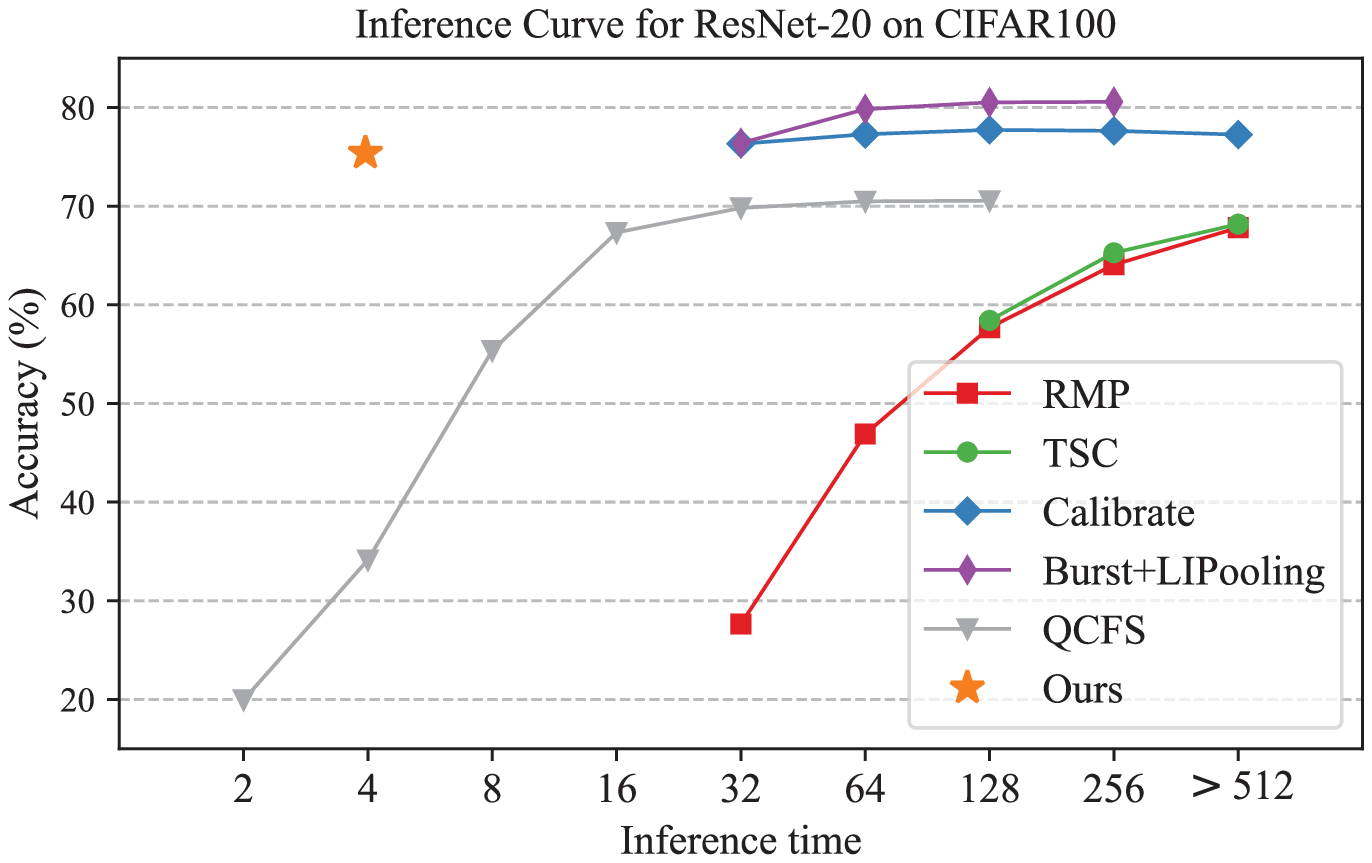}}  
\subfloat[ResNet-20 on Tiny-ImageNet]{
	\includegraphics[scale=0.42,trim=0 5 0 14,clip]{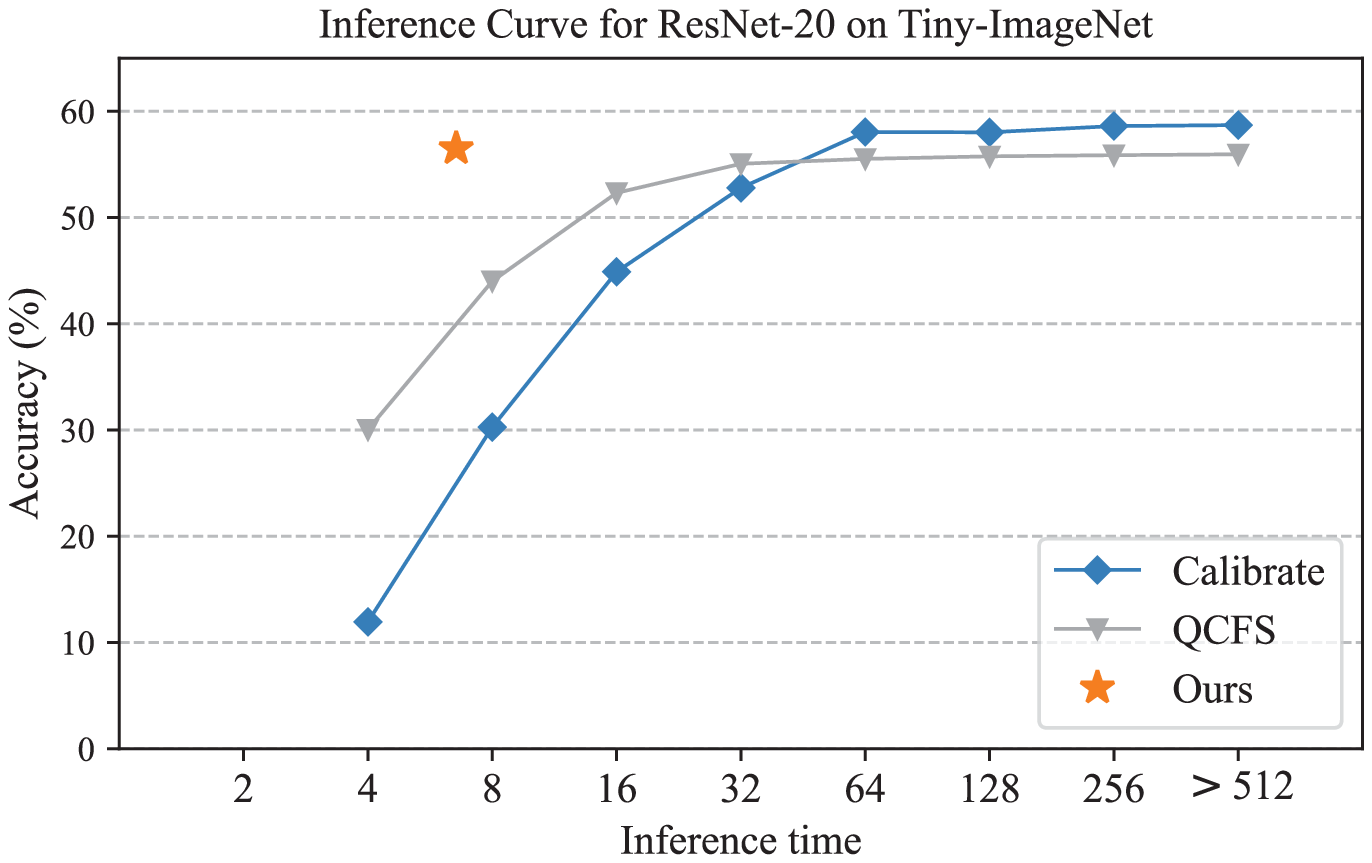}}
\vspace{-2mm}	
\caption{
Test accuracy as a function of inference time on the CIFAR-10, CIFAR-100, and Tiny-ImageNet datasets with \textbf{(a,b,c)} VGG-16 and \textbf{(d,e,f)} ResNet-20 architectures. Note that for the proposed hybrid coding approach, we only report the average inference time as the decision is made as soon as the first output spike is generated. }
\label{fig: overall_acc}
\end{figure*}

To achieve fast and accurate decision-making, the output layer is trained to make correct classification decisions based on the first output spike, which is referred to as TTFS learning hereafter. It is accomplished by carefully balancing two learning objectives. On the one hand, the weights of the target neuron should be adjusted to fire as early as possible to reduce inference latency. On the other hand, all the other neurons should fire later than the target one to ensure decision accuracy. These two learning objectives are optimized together by adjusting the weights of the output layer according to the following two loss functions, while keeping all other hidden layer weights fixed.

To achieve the first objective, we select the cross-entropy (CE) loss as the first loss function, which advances the first spike time of the target neuron by forcing its membrane potential to increase during training as:
\begin{equation}\label{loss1}
\mathcal{L}^L_1=-\ln{\frac{exp(\boldsymbol{v}^{L}_{target}(t_f))}{\sum_{i=1}^n exp(\boldsymbol{v}^{L}_{i}(t_f))}},
\end{equation}
where $\boldsymbol{v}^{L}_{i}(t_f)$ denotes the potential of the output neuron $i$ at the first spike time $t_f$, $n$ is the total number of categories. 

Given the fixed firing threshold and growing membrane potentials that result from minimizing $\mathcal{L}_1$, the decision time will be continuously advanced until every decision is made at the earliest possible timestep. This may, however, lead to an overfitting problem as the available input information is inadequately exploited. To address this issue, we introduce another loss, namely $\mathcal{L}_2$, to suspend incorrect decisions and allow more time to exploit input information. Specifically, long-term depression (LTD) will be triggered when any other neuron fires earlier than the target one as:
\begin{equation}\label{loss2}
\mathcal{L}^L_2=\sum^n_{i \neq target}{{\theta}(t_f)\cdot(\boldsymbol{v}_i^{L}(t_f)-V_{th}^{L})},
\end{equation}
where ${\theta}(t_f)$ is a scaling factor that is defined as:
\begin{equation}\label{factor}
{\theta}(t_f)=max\left(1-\frac{acc(t_f)}{acc(T)},0\right),
\end{equation}
where $acc(t_f)$ and $acc(T)$ correspond to the inference accuracy at the first spike time and the last timestep, respectively, whose value is updated after each evaluation. For incorrect decisions made too early, their firing time will be delayed to later timesteps depending on whether the accuracy can be improved if the decisions are made later, that is ${\theta}(t_f)>0$. 

The combination of the above two loss functions ensures those easy decisions can be made early while difficult decisions are postponed to later and more informative timesteps. The weight update of the output layer can be formulated as follows:
\begin{equation}\label{loss}
 \Delta \mathcal{W}^{L} \propto \frac{\partial \mathcal{L}^L}{\partial \mathcal{W}^{L}} = \frac{\partial \mathcal{L}^L}{\partial \boldsymbol{v}^{L}(t_f)} 
{K}^L(t_f),
\end{equation}
where the total loss $\mathcal{L}^L=\alpha \mathcal{L}^L_1+\beta \mathcal{L}^L_2$ is tuned by two hyperparameters $\alpha$ and $\beta$. {By moderating the ratio of $\beta/\alpha$, we strike a balance between the classification accuracy and the inference latency. Depending on the specific task requirements, a large $\beta/\alpha$ favors accuracy, and a small value minimizes inference latency (see more detailed analysis on the selection of $\beta$ and $\alpha$ in Supplementary Materials). In this work, we employ the metric {$\Delta \text{Acc}. \times t_{inf}$} as a means to trade-off between accuracy and inference latency. $\Delta \text{Acc}.$ denotes the accuracy drop from ANN to SNN, and $t_{inf}$, namely the average inference time, is equivalent to the mean first spike time $t_f$ across different samples.}

\begin{table*}[t]
\normalsize
\centering
\caption{Comparison of classification accuracy and inference time on the CIFAR-10, CIFAR-100, and Tiny-ImangeNet datasets.
$\Delta \text{Acc.}= \text{SNN Acc.} - \text{ANN Acc.}$ }
\vspace{-2mm}
\label{tab:comp_overall}
\resizebox{0.8\textwidth}{!}{%
\begin{tabular}{l l l c c c c}
\hline
\hline
\textbf{Dataset}  & \textbf{Method}     & \textbf{Architecture} &\textbf{ANN Acc. (\%)} &\textbf{SNN Acc. (\%)} &\textbf{$\Delta$ Acc. (\%)} &\textbf{Avg. Inference Time}\\ 
\hline
\multirow{17}{*}{CIFAR-10}      & Burst Spikes \cite{2019fast}          & VGG-16        & 91.41    & 91.41     & 0           & 793 \\
      & RMP \cite{han2020rmp}                 & VGG-16        & 93.63    & 93.04     & -0.59    & 256 \\
      & TSC \cite{han2020deep}                & VGG-16        & 93.63    & 93.27     & -0.36    & 128 \\
      & Calibration \cite{Calibration}        & VGG-16        & 95.72    & 95.14     & -0.58     & 64 \\
      & Burst+LIPooling \cite{li2022efficient}& VGG-16        & 95.74    & 95.58     & -0.16     & 32 \\
      & QCFS \cite{QCFS}                      & VGG-16        & 95.52    & 95.40     & -0.12     & 16 \\
      & Diet-SNN \cite{rathi2021diet}         & VGG-16        & 93.72    & 93.44     & -0.28     & 10 \\
      & \textbf{Ours (Hybrid Coding)}         & \textbf{VGG-16}        & \textbf{95.84}    & \textbf{95.66}     & \textbf{-0.17}     & \textbf{4.87} \\

      & RMP \cite{han2020rmp}                 & ResNet-20     & 91.47    & 91.36     & -0.11     & 2048 \\
      & TSC \cite{han2020deep}                & ResNet-20     & 91.47    & 91.42     & -0.05     & 2048 \\
      & Calibration \cite{Calibration}        & ResNet-20     & 95.46    & 94.78     & -0.68     & 32 \\
      & Burst+LIPooling \cite{li2022efficient}& ResNet-20     & 96.56    & 96.11     & -0.45     & 32 \\
      & QCFS \cite{QCFS}                      & ResNet-20     & 91.77    & 91.62     & -0.15     & 16 \\
      & Diet-SNN \cite{rathi2021diet}         & ResNet-20     & 92.79    & 92.54     & -          & 10 \\
      & STBP-tdBN \cite{zheng2021going}       & ResNet-19$^*$     & -        & 93.16     & -          & 6 \\
      & TET \cite{deng2022temporal}           & ResNet-19$^*$     & -         & 94.50     & -          & 6 \\
      & \textbf{Ours (Hybrid Coding)}           & \textbf{ResNet-20}    & \textbf{95.45}    & \textbf{95.24}     & \textbf{-0.21}     & \textbf{5.04} \\
\hline
\multirow{16}{*}{CIFAR-100}      & Burst Spikes \cite{2019fast}          & VGG-16        & 68.77    & 68.69     & -0.08           & 3000 \\
      & RMP \cite{han2020rmp}                 & VGG-16        & 71.22   & 70.93     & -0.29    & 2048 \\
      & TSC \cite{han2020deep}                & VGG-16        & 71.22    & 70.97     & -0.25    & 2048 \\
      & Calibration \cite{Calibration}        & VGG-16        & 77.89    & 76.64     & -1.25     & 64 \\
      & Burst+LIPooling \cite{li2022efficient}& VGG-16        & 78.49    & 78.26     & -0.23     & 64 \\
      & QCFS \cite{QCFS}                      & VGG-16        & 76.28    & 76.24     & -0.04     & 16 \\
      & Diet-SNN \cite{rathi2021diet}         & VGG-16        & 71.82    & 69.67     & -2.15     & 5 \\
      & \textbf{Ours (Hybrid Coding)}         & \textbf{VGG-16}        & \textbf{77.15}    & \textbf{75.75}     & \textbf{-1.40}     & \textbf{6.88} \\
      & RMP \cite{han2020rmp}                 & ResNet-20     & 68.72    & 67.82     & -0.90     & 2048 \\
      & TSC \cite{han2020deep}                & ResNet-20     & 68.72    & 68.18     & -0.54     & 2048 \\
      & Calibration \cite{Calibration}        & ResNet-20     & 77.16    & 76.32     & -0.84     & 32 \\
      & Burst+LIPooling \cite{li2022efficient}& ResNet-20     & 80.69    & 79.83     & -0.86     & 64 \\
      & QCFS \cite{QCFS}                      & ResNet-20     & 69.94    & 69.82     & -0.12     & 32 \\
      & Diet-SNN \cite{rathi2021diet}         & ResNet-20     & 64.64    & 64.07     & -0.57     & 5 \\
      & TET \cite{deng2022temporal}           & ResNet-19$^*$     & -         & 74.72     & -          & 6 \\      
      & \textbf{Ours (Hybrid Coding)}         & \textbf{ResNet-20}    & \textbf{76.56}    & \textbf{75.39}     & \textbf{-1.17}     & \textbf{3.94} \\
\hline
\multirow{8}{*}{Tiny-ImageNet}      & Spike-thrift \cite{kundu2021spike}    & VGG-16        & 56.56    & 51.92     & -4.64     & 150 \\
      & Calibration \cite{Calibration}        & VGG-16$^+$        & 60.95    & 53.96     & -6.99     & 32 \\
      & Burst+LIPooling \cite{li2022efficient}& VGG-16$^+$        & 61.80    & 55.56     & -6.24     & 64 \\
      & QCFS \cite{QCFS}                      & VGG-16$^+$        & 54.91    & 54.92     & 0.01      & 16 \\
      & \textbf{Ours (Hybrid Coding)}         & \textbf{VGG-16}          & \textbf{58.75}    & \textbf{56.95}     & \textbf{-1.80}     & \textbf{7.11}\\
      & Calibration \cite{Calibration}        & ResNet-20$^+$     & 58.29    & 52.79     & -5.5      & 32 \\
      & QCFS \cite{QCFS}                      & ResNet-20$^+$     & 55.81    & 52.32     & -3.49     & 16 \\
      & \textbf{Ours (Hybrid Coding)}         & \textbf{ResNet-20}     & \textbf{58.34}    & \textbf{56.50}     & \textbf{-1.84}     & \textbf{6.54} \\
\hline
\hline
\multicolumn{7}{l}{* Only the results of ResNet-19 are reported in these works, which have comparable model capacity to other ResNet-20 models.}\\
\multicolumn{7}{l}{+ Our reproduced results using publicly available codes.}
\end{tabular}}
\vspace{-3mm}
\end{table*}

\section{Accurate and Rapid Image classification with Hybrid coding}
\label{sec:exp}
In this section, we compare the proposed hybrid coding approach to other SOTA SNN implementations on the CIFAR-10 \cite{krizhevsky2009learning}, CIFAR-100 \cite{krizhevsky2009learning}, and Tiny-ImageNet \cite{wu2017tiny} datasets, in terms of two primary performance indicators of image classification tasks: classification accuracy and inference latency \cite{li2022image}. For all datasets, we utilize both VGG-16 \cite{simonyan2014very} and ResNet-20 \cite{he2016deep} network structures. Subsequently, we validate the effectiveness and convergence speed of our proposed layer-wise learning method. Lastly, we demonstrate the proposed approach can achieve SOTA classification accuracy while significantly reducing the computational cost. For simplicity, the direct, rate, burst, phase, and TTFS coding schemes are denoted as D, R, B, P, and T, respectively, in the following sections. {The source codes are available at \href{https://github.com/xychen-comp/Hybrid-Coding-SNN}{https://github.com/xychen-comp/Hybrid-Coding-SNN}}.

\subsection{Superior Image Classification Performance}
In Table \ref{tab:comp_overall}, we report our results and compare them against other SOTA SNN models. To ensure a fair comparison, we select competitive SNN benchmarks based on their conversion errors for network conversion methods, or classification accuracy for direct training methods. Overall, as the inference curves illustrated in Fig. \ref{fig: overall_acc}, the proposed hybrid coding approach attains competitive classification accuracy while significantly reducing inference time across all model architectures and datasets.

For the CIFAR-10 dataset, our VGG-16 model requires only 4.87 timesteps on average to reach 95.66\% classification accuracy. Whereas the QCFS \cite{QCFS}, considered the best among all ANN-to-SNN conversion methods, can only achieve a comparable accuracy with 16 timesteps. Similar conclusions can be drawn from the converted ResNet-20 models. Furthermore, our ResNet-20 model outperforms other directly trained models, such as STBP-tdBN \cite{zheng2021going} and TET \cite{deng2022temporal}. 

For the more challenging CIFAR-100 dataset, our model reaches a satisfactory accuracy of 75.75\% and 75.39\% with average inference timesteps of 6.88 and 3.94 for VGG-16 and ResNet-20, respectively. As presented in Fig. \ref{fig: overall_acc}(b, e), although the classification accuracy of our models is slightly worse than Calibration \cite{Calibration}, Burst+LIPooling \cite{li2022efficient}, and QCFS models, it is worth noting that their results are achieved with at least 16 timesteps. Consequently, our models are more favorable in terms of inference time and energy consumption. Additionally, we evaluate our method on a large-scale Tiny-ImageNet dataset. Due to a lack of benchmarks on this dataset, we reproduce the results for Calibration, Burst+LIPooling, and QCFS methods using their publicly available codes. As shown in Fig. \ref{fig: overall_acc}(c, f), it is obvious that our proposed models are capable of achieving superior results with a shorter inference time.

\begin{figure}[tb]
\centering
{\includegraphics[scale=0.65, trim=0 0 -10 0,clip]{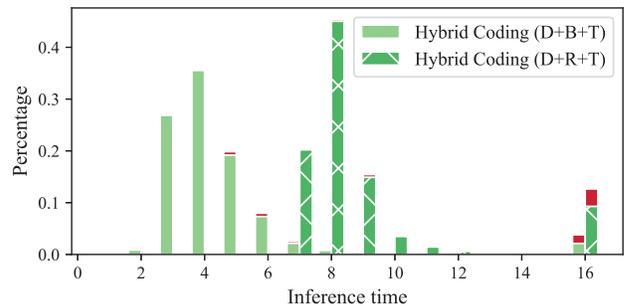}}\label{fig: distribution}\vspace{-2mm}
\caption{{Decision time distribution for the proposed hybrid coding models (D+B+T and D+R+T). 
The decision times for each sample vary according to the TTFS coding mechanism. \textbf{Green Bar:} the proportion of correct decisions. \textbf{Red Bar}: the proportion of misclassified samples.} }
\label{fig:distri}
\end{figure}

{To elucidate how early decision-making is accomplished with the proposed hybrid coding SNN, we present in Fig. \ref{fig:distri} the distribution of decision time across two of our hybrid coding models, D+B+T and D+R+T. 
While the other SOTA works listed in Table \ref{tab:comp_overall} employ direct coding in the output layer, resulting in a constant decision time of $T$. In contrast, our hybrid coding models exhibit varying decision times governed by the decision-making mechanism of TTFS coding. Furthermore, it is evident that the first spike time is densely distributed around 4 for D+B+T and 8 for D+R+T models, respectively. This result substantiates that burst coding facilitates faster information transmission across hidden layers, which forms the basis for the early decisions made by the TTFS classifier. Additionally, it is noteworthy that decisions made at earlier timesteps exhibit higher accuracy than those made later, as indicated by the red bars. This observation suggests early decisions typically correspond to simpler samples, and TTFS coding is capable of identifying when sufficient evidence has been accumulated for different samples. Compared with the direct decoding that makes all decisions at the final timestep, TTFS coding can significantly reduce average inference time yet achieve high classification accuracy. }

\begin{figure*}[t]
\centering
\includegraphics[scale=0.55, trim=15 85 0 80,clip]{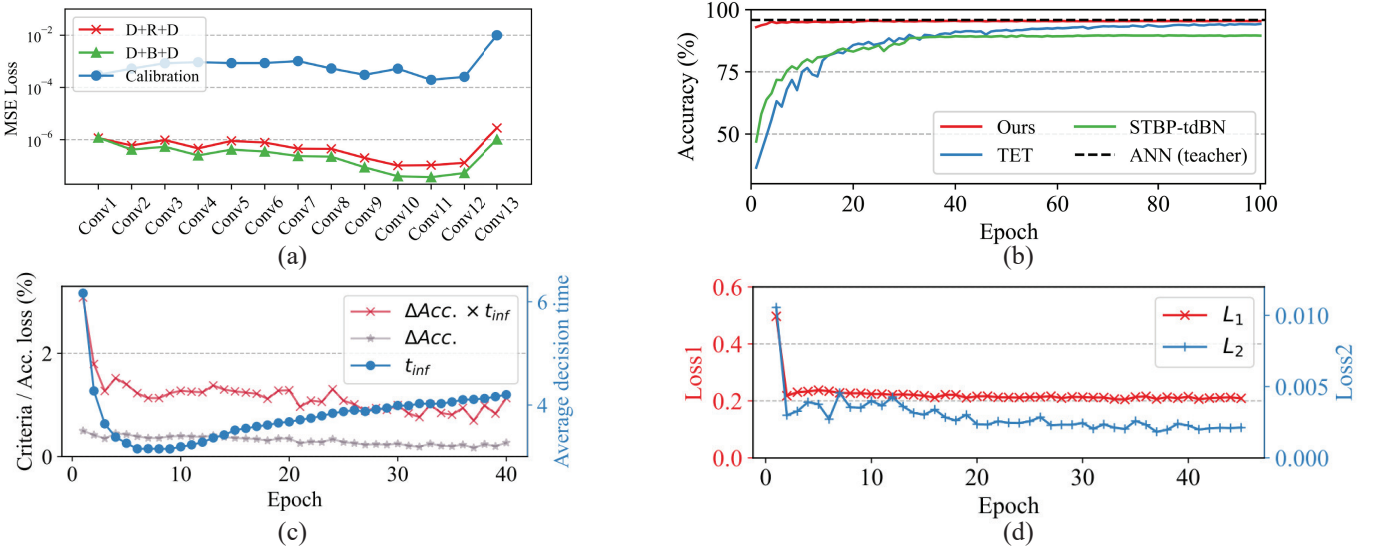}
\vspace{-4mm}
\caption{{\textbf{(a)} {Performance validation of our proposed LTL learning method in achieving desired burst coding across hidden layers. We compare our method with the Calibration method, which bears the closest similarity to ours in terms of the underlying mechanism. \textbf{(b)} Learning curves comparison among our layer-wise training method LTL, direct training method STBP-tdBN, and TET, highlighting the faster convergence of our LTL training method.} \textbf{(c)} Illustration of learning dynamics during TTFS learning. $t_{inf}$ refers to the average inference time, $\Delta \text{Acc}.$ is the accuracy drop from baseline ANN to SNN, and $\Delta \text{Acc}. \times t_{inf}$ is the evaluation criterion used in our study to assess the overall model performance. \textbf{(d)} The evolution of the two training losses used during TTFS learning, which govern the learning dynamics observed in (c).}}
\label{fig: 4in1}
\end{figure*}

\subsection{Effective and Rapid Training Convergence with Layer-wise Learning Method}
\label{sec:effective_training}

To further explore the effectiveness and training convergence speed of the proposed layer-wise learning method, we conduct a series of studies in this section. All experiments are performed on the CIFAR-10 dataset using VGG-16 architecture. 

Firstly, to assess the effectiveness of the LTL rule in producing the desired hidden layer representations as specified by the neural coding scheme, we systematically compare the performance of the Calibration model \cite{Calibration}, the LTL-trained D+R+D and D+B+D models. The reason for choosing the Calibration method for comparison is that, among all the considered SOTA methods, this particular approach bears the closest resemblance to our proposed LTL learning method in terms of its underlying mechanism. The LTL-trained D+R+D and D+B+D models exhibit minimal accuracy drops of 0.56\% and 0.31\%, respectively, from the teacher ANN. This suggests the LTL rule is highly effective in producing both rate- and burst-based neural representations. In the same vein of research, the Calibration method \cite{Calibration} is proposed to fine-tune the hidden layer representations in a layer-wise manner, following the principle of rate coding. However, this method neglects the temporal dynamics of spike trains and only conducts training at the spike-train level. Consequently, their model employing D+R+D coding encounters a larger accuracy drop of 1.21\%. 

To analyze the underlying reason for this performance improvement, we calculate the normalized MSE losses (i.e., Eq. (\ref{eq: LTL_loss})) between the convolutional layers of teacher ANN and student SNN, the results are shown in Fig. \ref{fig: 4in1}(a). As expected, the LTL-trained D+R+D model achieves significantly lower MSE losses than those of the Calibration model. This finding suggests that fine-tuning the temporal dynamics of each hidden layer is crucial for achieving the desired neural representations. Furthermore, the LTL-trained D+B+D model outperforms the D+R+D model slightly, indicating burst coding can better approximate the neural representations of teacher ANN than rate coding.

Additionally, by leveraging the knowledge from teacher ANN, the layer-wise LTL rule can achieve faster training convergence than other end-to-end training rules. {To illustrate this, we compare the learning curves of the LTL rule with the STBP-tdBN method \cite{zheng2021going}, and the TET rule \cite{deng2022temporal}. These two methods are end-to-end SNN training methods that are directly comparable to our method, while other methods mentioned employ ANN-to-SNN conversion methods to obtain network parameters from a pre-trained ANN. As depicted in Fig. \ref{fig: 4in1}(b), our model reaches the accuracy of 95.19\% rapidly within only 5 training epochs, whereas {the STBP-tdBN model converges after 50 epochs and} the TET model does not converge even at epoch 100. Beyond the early decision-making capability as discussed earlier, our hybrid neural coding and learning framework also significantly reduces the cost of training and deployment of SNNs.}

In general, the quality of the decision improves when a longer inference time is involved in an SNN. To illuminate how the proposed TTFS learning rule effectively reduces the latency without compromising classification accuracy, we plot the learning curves of the TTFS classifier in Fig. \ref{fig: 4in1}(c, d). Both accuracy and inference latency are objectives to be optimized during training. Specifically, we employ $\Delta \text{Acc}. \times t_{inf}$ as the evaluation criteria.
As shown in Fig. \ref{fig: 4in1}(c), the average decision time is sharply advanced at the beginning, accompanied by a degree of accuracy loss due to overfitting. The learning dynamics during this period are mainly governed by the long-term potentiation (LTP) effect of loss function $\mathcal{L}_1$ as described in Fig. \ref{fig: 4in1}(d). Fortunately, the training transits to be dominated by $\mathcal{L}_2$ after the convergence of $\mathcal{L}_1$, which improves the classification accuracy by slightly delaying wrong decisions, allowing more input evidence to accumulate. The learning dynamics at these two stages perfectly align with the two objectives specified in the TTFS learning rule, thereby ensuring an early yet accurate decision is made.

\begin{table*}[htb]
\normalsize
\centering
\caption{Comparison of energy efficiency with other SNN implementations when achieving a comparable classification accuracy. \textbf{$\Delta\text{Acc.}$}: the accuracy difference between equal-structured ANN and SNN. \textbf{Ratio}: SynOps(SNN)/SynOps(ANN). \textbf{Energy Saving}: Estimated energy saving of SNN against ANN using 32-bit float operations (32-bit integer operations). }
\label{tab:synops}
\resizebox{0.8\textwidth}{!}{%
\begin{tabular}{l c c c c c }
\hline
\hline
\textbf{Model}  &\textbf{SNN Acc. (\%)}  &\textbf{$\Delta$Acc. (\%)}  & \textbf{Avg. Inference Time}   &\textbf{Ratio}  &\textbf{Energy Saving}\\ 
\hline
Calibration \cite{Calibration}            & 95.04     & -0.65            & 64    & 2.10   & 2.43 (15.24) \\
QCFS \cite{QCFS}                        & 93.43     & -0.15            & 16    & 1.91   & 2.68 (11.94)\\
Burst+LIPooling \cite{li2022efficient}  & 93.71     & -0.40             & 64    & 4.81   & 1.06 (6.65)\\
TET \cite{deng2022temporal}  & 94.94     & -             & 6    & 2.06   & 2.48 (15.53)\\
\textbf{Ours (Hybrid Coding)}                     & \textbf{95.66}     & \textbf{-0.17}            & \textbf{4.87}  & \textbf{0.49}  & \textbf{10.43 (65.31)}\\
\hline
\hline
\multicolumn{5}{l}{}
\end{tabular}}
\vspace{-4mm}
\end{table*}

\subsection{Superior Energy Efficiency with Hybrid Coding}
\label{sec:energy}
In the previous sections, we demonstrated that the proposed hybrid neural coding and learning framework can achieve rapid decision-making. Here, we analyze whether this attribute can further contribute to energy savings. 
We adhere to a common practice to assess the energy consumption of SNNs in this work, which was initially proposed by IBM TrueNorth neuromorphic chip in their seminal Science paper \cite{merolla2014million}. It has since become the prevailing standard in the neuromorphic computing community for benchmarking the energy efficiency of different SNN and ANN models \cite{yin2021accurate,yao2023attention,wu2021progressive}. 
This method focuses on quantifying the energy consumption associated with the basic operations in SNNs and ANNs, i.e., synaptic operations (SynOps). It provides a convenient and reliable way to estimate and compare energy consumption across different models, thus enabling equitable energy performance evaluations. 
To this end, we calculate the SynOps ratio between equal-structured ANN and SNN by multiply-and-accumulate (MAC) and accumulate (AC) operations, respectively. The total SynOps required in ANN models can be calculated as: 
\begin{equation} 
	\text{SynOps}(\text{ANN}) = \sum\limits_l^Lf_{in}^lN^l,
\label{eq: synops ann}
\end{equation}
where $L$ is the total number of layers, $f^l_{in}$ denotes the fan-in connections to layer $l$ from the previous layer, and $N^l$ is the total number of neurons in layer $l$. As SNN models have an additional temporal dimension and only operate when spikes occur, their total SynOps calculation is different from their ANN counterparts. Hence, the total SynOps in SNN models are formulated as below: 
\begin{equation}
    \text{SynOps}(\text{SNN}) = \sum\limits_t^T \sum\limits_l^{L-1} \sum\limits_j^{N^l} f_{out,j}^l\boldsymbol{s}_j^l(t),
	\label{eq: synops snn}
\end{equation}
where $T$ is the total inference time, $f^l_{out}$ represents the fan-out connections from neuron $j$ in layer $l$ to the next layer, and $\boldsymbol{s}^l_j(t)$ denotes the spike count of neuron $j$.

To facilitate a fair comparison with other SOTA SNN models, we find recent works whose codes are publicly available and reproduce their results on the CIFAR-10 dataset using VGG-16 architecture. Particularly, we report the SynOps ratios at the timestep when a comparable classification accuracy has been achieved. 
As the results summarised in Table \ref{tab:synops}, our model achieves the lowest SynOps ratio of 0.49, which is at least 3.90$\times$ lower than the other SOTA models. Notably, the total SynOps required by our model is much lower than that of the conversion method QCFS, which achieves a SynOps ratio of 1.91 at $T=16$. The observed discrepancy in energy consumption further validates our initial findings, suggesting that the reduction in inference latency is not primarily due to the compressed time window offered by burst coding. Instead, burst coding allows important information to be transmitted earlier, allowing early and efficient decisions made by TTFS coding. Surprisingly, the SynOps ratio of the TET directly trained model is significantly higher than ours, even at a comparable inference time. We attribute this to the activity regularization loss introduced in TET training. Specifically, this loss term promotes the stable generation of spikes over time. Once removing this loss term, the models tend to spike more frequently during the final few time steps, which in turn reduces the overall firing rate. 

Empirical evidence from a recent study \cite{han2015learning} on 45 nm CMOS technology reveals that the 32-bit float MAC and AC operations consume 4.6 pJ and 0.9 pJ energy, respectively. As such, our SNN model can yield an order of magnitude energy saving over its ANN counterpart. The saving can be further boosted to 65.31$\times$ when the cheaper 32-bit integer operations are used (3.1 pJ and 0.1 pJ for corresponding MAC and AC operations).

\section{Efficient and Robust Sound Localization with Hybrid Coding}
In this section, we evaluate the proposed hybrid neural coding and learning framework on the SLoClas dataset \cite{qian2021sloclas}, which is designed for real-world sound localization and classification tasks. Particularly, we comprehensively compare the hybrid coding SNN against other SOTA models focusing on localization precision, inference latency, energy consumption, and noise robustness. Moreover, several ablation studies are carried out to gain insight into the impact of different hybrid neural coding combinations on the overall performance. Detailed configurations of data preprocessing, MTPC front-end, and SNN architecture are summarized in Supplementary Materials. 

\subsection{Precise, Rapid, and Robust Sound Localization}
\label{sec:precise}

\begin{table*}[t]
\normalsize
\centering
\caption{Comparison of sound localization accuracy, MAE, and inference time on the SLoClas dataset.}
\vspace{-2mm}
\label{tab:SL_results}
\resizebox{0.8\textwidth}{!}{%
\begin{tabular}{l l c c c c c}
\hline
\hline
\textbf{Dataset} & \textbf{Method}  & \textbf{SNN} & \textbf{Parameters} & \textbf{MAE ($^\circ$)} & \textbf{Accuracy (\%)} & \textbf{Avg. Inference Time} \\ 
\hline
\multirow{7}{*}{SLoClas}     &GCC-PHAT-CNN \cite{qian2021sloclas} &No  & 4.17M   & 4.39          & 86.94    & - \\
    &SSLNN \cite{he18b_interspeech}$^*$ &No  & 1.60M &3.13& 85.13& -\\
    &SELDnet \cite{adavanne2018sound}$^*$ &No  & 1.68M & 1.78 & 88.24 & -\\
    & EINV2 \cite{EINV2}$^*$ &No  & 1.63M & 0.98 & 94.64& -\\
    &SRP-DNN \cite{yang2022srpdnn}$^*$&No  & 1.64M &0.96 & 94.12 & -\\
    &FN-SSL \cite{wang2023fnssl}$^*$&No  & 1.68M &0.63 &95.40 & -\\

    & MTPC-CSNN \cite{Multitone}$^*$ &Yes  & 1.61M & 1.23 &   93.95    & 4 \\
    & MTPC-CSNN \cite{Multitone}$^*$ &Yes  & 1.61M & 1.02 &   94.72    & 8 \\
    &MTPC-RSNN \cite{Multitone}$^*$ &Yes  & 1.67M & 1.48 &   94.30    & 51 \\
      
      & \textbf{Ours (Hybrid Coding)}  &Yes   & 1.61M   & \textbf{0.60}     & \textbf{95.61}     & \textbf{4.37}\\
\hline
\hline
\multicolumn{7}{l}{* \quad Our reproduced results are based on publicly available source code. }\\
\end{tabular}}
\end{table*}

To provide a comprehensive assessment of the localization precision for our proposed model and other SOTA methods, we report both the mean absolute error (MAE) and the classification accuracy defined as follows \cite{qian2021sloclas}:
\begin{equation}
\begin{aligned}
\text{MAE}\ (^\circ) &=\frac{1}{N_t} \sum_{i=1}^{N_t}|\hat{\theta}_{i}-\theta_{i}|, \\
\text{Acc.}\ (\%) &=\frac{1}{N_t} \sum_{i=1}^{N_t}(|\hat{\theta}_{i}-\theta_{i}|<\eta),
\end{aligned}
\label{mae_acc}
\end{equation}
where $\hat{\theta}_{i}$ and $\theta_{i}$ are the estimated and the ground truth azimuth angle of sample $i$. $\eta$ denotes the acceptable absolute error in determining whether a sample is correctly classified. $N_t$ is the total number of test samples. Given the raw audio is recorded at a resolution of 5$^\circ$, we maintain consistency by setting $\eta$ to 2.5$^\circ$. Consequently, $\hat{\theta}_{i}$ is rounded to the nearest 5-degree increment when calculating classification accuracy.

As presented in Table \ref{tab:SL_results}, our hybrid coding SNN achieves an MAE of 0.60$^\circ$ with an average inference time of only 4.37 timesteps, {outperforming all other SOTA SNN models while requiring comparable or shorter inference time. The localization precision is also competitive with other recently introduced ANN models.}

\begin{figure}[!t]
\centering
\includegraphics[scale=0.62,trim= 0 0 0 0, clip]{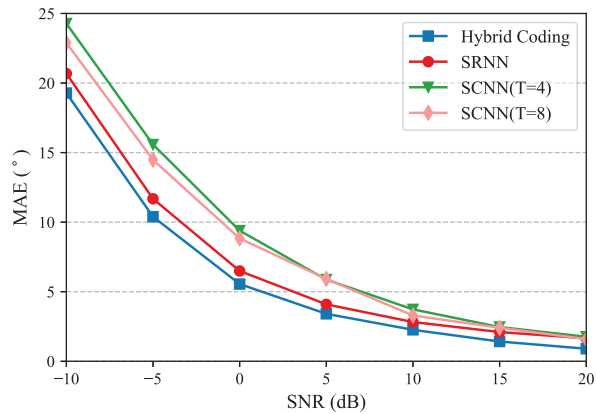}
\caption{Comparison of localization precision under different SNRs among different SNN models.}
\label{fig: NOISE}
\end{figure}

Furthermore, the noise robustness of the proposed model is evaluated against other SNN models. The directional background noise fragments provided in the SLoClas dataset are randomly selected and added to the test audio clips at different signal-to-noise ratios (SNRs). As demonstrated in Fig. \ref{fig: NOISE}, our hybrid coding SNN consistently demonstrates higher robustness against background noise, regardless of the levels of noise present.

\subsection{Ablation Study of Hybrid Coding Design}
\label{sec:ablation}
To systematically investigate the impact of various hybrid coding combinations for the sound localization task, we have conducted a series of ablation studies. Specifically, we thoroughly analyze the four most promising combinations: P+R+D, P+B+D, P+R+T, and P+B+T.

\subsubsection{Localization Precision, Latency, and Energy Consumption}

\begin{table}[t]
\normalsize
\centering
\caption{ Comparison of MAE, inference time, and SynOps among different neural coding combinations.}
\vspace{-2mm}
\label{tab:abla}
\resizebox{0.45\textwidth}{!}{%
\begin{tabular}{c c c c c}
\hline
\hline
\textbf{Model} & \textbf{MAE ($^\circ$)} & \makecell[c]{\textbf{Avg. Inference} \\ \textbf{Time}}  & 
\makecell[c]{\textbf{SynOps} \\ \textbf{(Millions)}} \\
\hline
ANN & 1.12 & - & 5.15 \\ 
\hline
 P+R+D & 0.83 & 16 & 2.06\\
 \hline
 P+B+D & 0.61 & 16 &  10.70\\
 \hline
 P+R+T & 0.82 & 7.62 & 0.94\\
 \hline
 P+B+T & 0.60 & 4.37 & 2.86\\
\hline
\hline
\end{tabular}}
\end{table}

\begin{figure*}[t]
\centering
\subfloat[]{
	\includegraphics[scale=0.43, trim= 0 0 -10 0, clip]{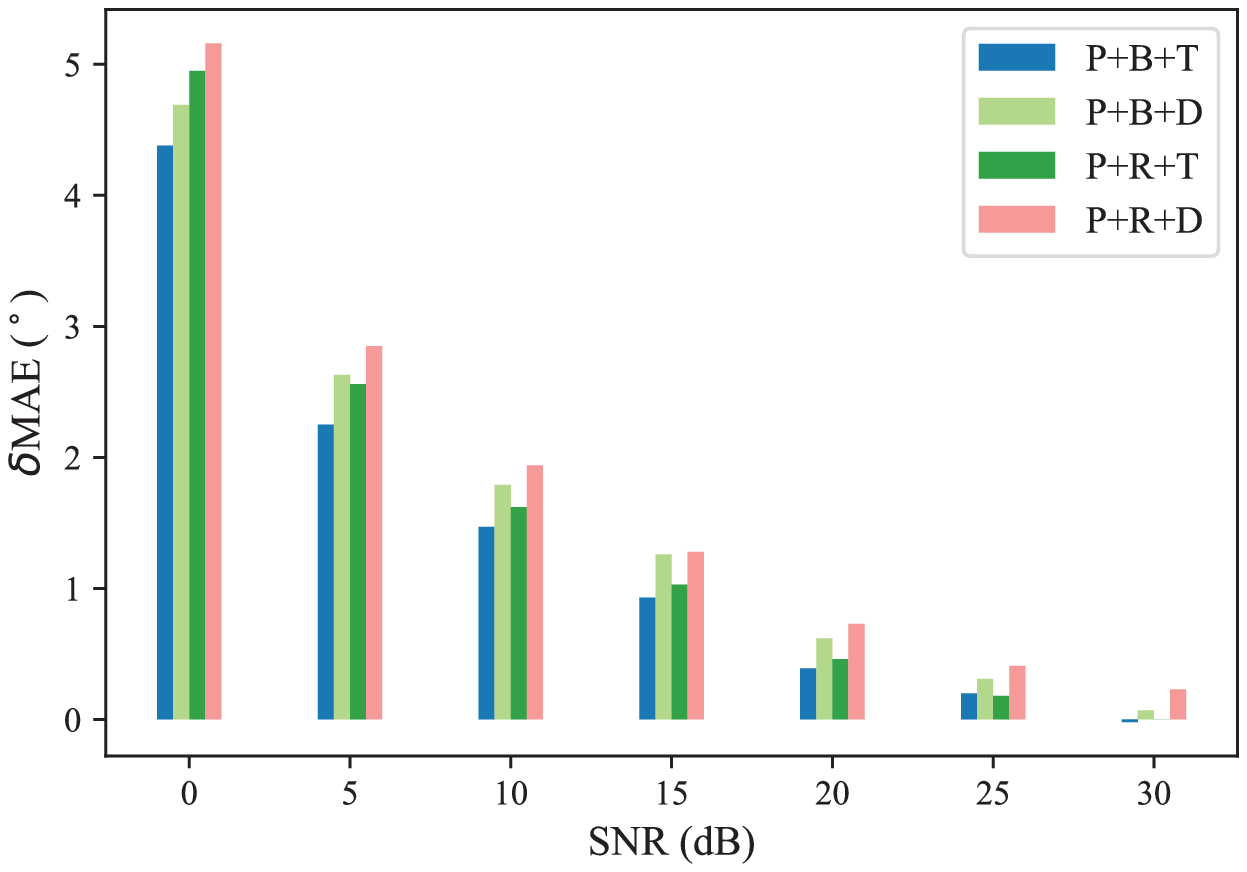}}\hspace{0em}
\subfloat[]{
	\includegraphics[scale=0.43, trim= 0 0 -10 0, clip]{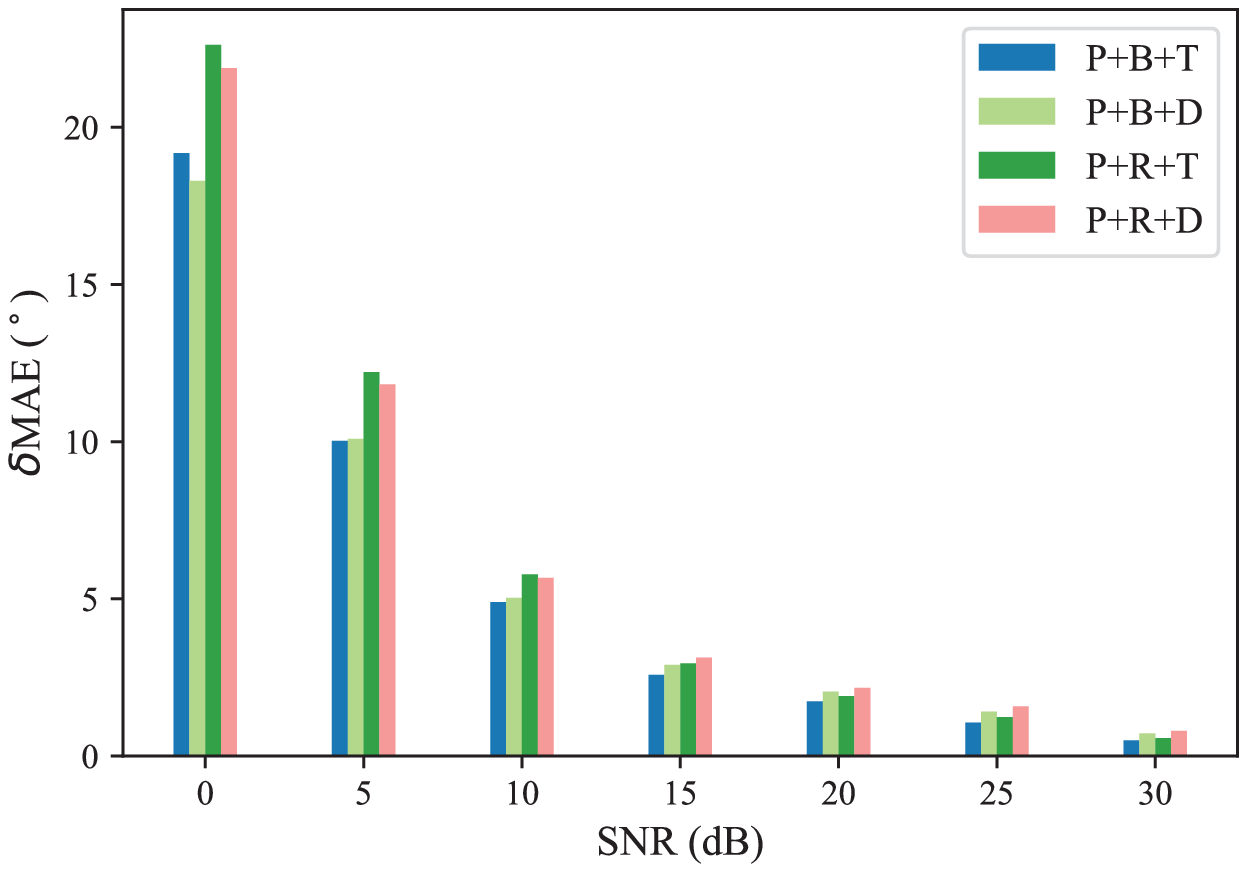}}\hspace{0em}
 \subfloat[]{
	\includegraphics[scale=0.43,trim= 0 0 0 0, clip]{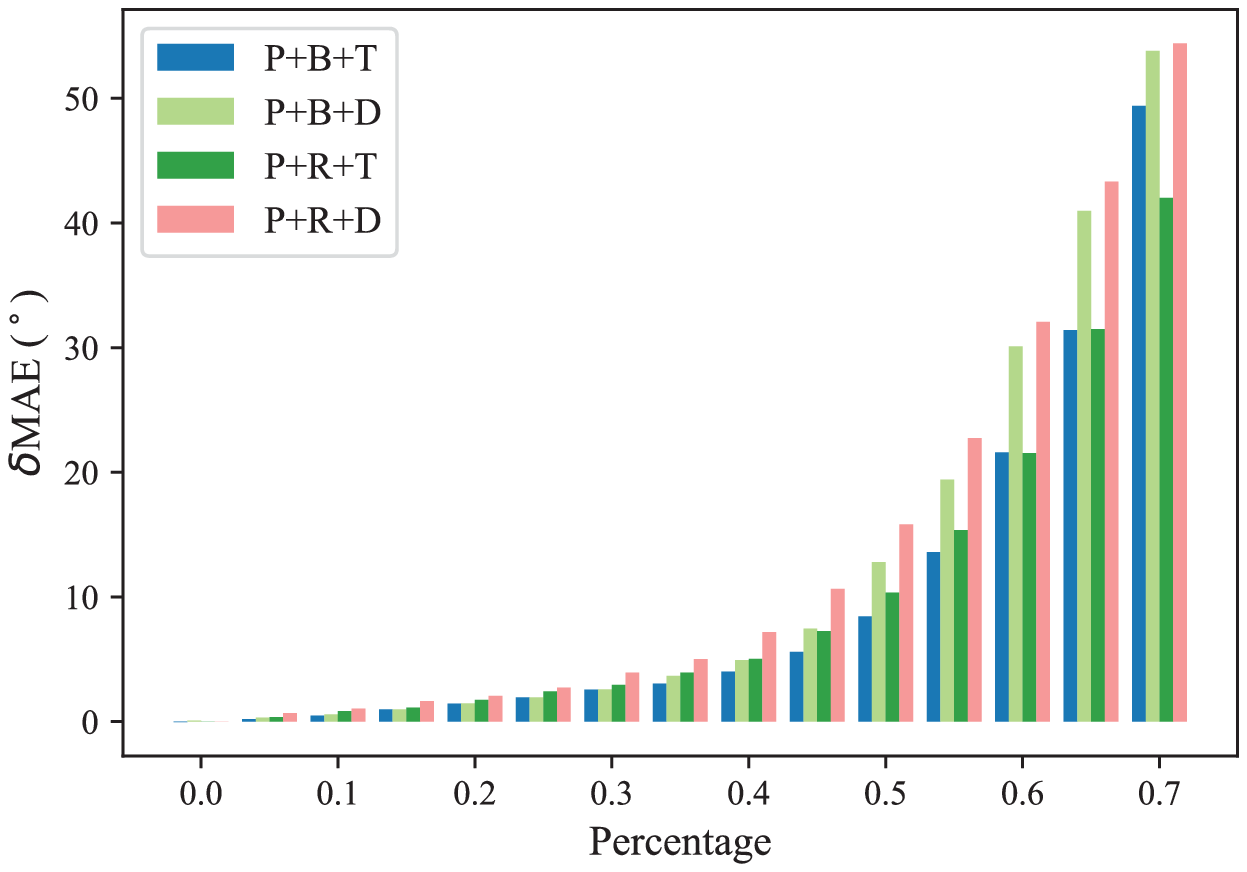}}\hspace{0em}
 \vspace{-2mm}
\caption{The sound localization precision comparison under different noise conditions: \textbf{(a)} babble noise, \textbf{(b)} factory noise, \textbf{(c)} neuronal noise. \textbf{$\delta\text{MAEs}$} is the MAE discrepancy between noisy and clean conditions.}
\label{fig: ablation}
\end{figure*}

We first evaluate the MAE, inference time, and SynOps of different hybrid coding designs, as demonstrated in Table \ref{tab:abla}.
To investigate the impact of the output decoding scheme on network performance, we first conduct a comparative analysis between P+B+T and P+B+D, as well as between P+R+T and P+R+D. Obviously, direct decoding and TTFS coding yield nearly identical MAEs, whereas TTFS coding only requires about one-third to one-half of the average inference time. These results further support our claim that TTFS coding can lead to reduced inference latency and energy consumption while maintaining high precision. Intriguingly, the ratios of energy efficiency and inference time between the two output coding schemes exhibit consistent value, irrespective of what coding schemes are used in the hidden layers. This observation implies that the reduced inference time plays a pivotal role in total energy saving. 

Furthermore, the results also offer valuable insights into the impact of hidden layer coding schemes. Burst coding outperforms rate coding by reducing 0.22$^\circ$ MAE while accelerating decision-making by 3.25 timesteps. However, burst coding may result in increased spike firing rates, in turn leading to higher energy consumption. To optimize the network performance, the choice between burst coding and rate coding is contingent upon the specific requirements of the task at hand. For instance, if the priority is given to energy efficiency, the P+R+T design could be the preferred choice. Alternatively, if minimizing MAE and latency are the primary concerns, the P+B+T design becomes an unquestionably superior choice. Notably, even though it demands more energy than the P+R+T design, it still only requires half the energy consumption of a competitive ANN model, striking an optimal balance between precision, latency, and energy consumption.

\subsubsection{Environmental Noise}
In order to investigate the impact of hybrid coding designs on noise robustness, we generate noise-corrupted test samples using two distinct forms of environmental noise, namely factory noise and babble noise. All models are trained on clean train samples and subsequently evaluated on noise-corrupted test samples with varying SNRs. The results of different hybrid coding designs are presented in Fig. \ref{fig: ablation}(a, b). In general, the P+B+T design demonstrates the highest noise robustness under most conditions. This notable advancement can be attributed to the fact that TTFS coding is less susceptible to cumulative noises, as it only requires about one-third of the time for decision-making. However, TTFS coding may also yield adverse effects when environmental noise significantly impairs information integrity, as it can only leverage limited evidence due to the reduction of inference time. Furthermore, our results suggest that hybrid coding SNNs consistently perform better under babble noise than under factory noise, owing to the fact that the latter has larger frequency variations.

\subsubsection{Neuronal Noise}
Furthermore, we analyze the robustness of hybrid coding designs against neuronal noise, which is known to be ubiquitous in human neural systems. Investigating the network performance under neuronal noise provides an opportunity to identify the most promising design that closely mimics the human auditory systems. The neuronal noise is simulated by introducing spike deletion operations, which means in each timestep, a certain fraction of hidden neurons fail to generate output spikes at a given probability. The results are visualized as a function of deletion rate in Fig. \ref{fig: ablation}(c). It indicates that TTFS coding is more reliable than direct decoding under neuronal noise conditions, which is consistent with our observations on environmental noises. Interestingly, when a large percentage of information fails to transfer ($>$60\%), the P+R+T design surpasses the P+B+T design in performance. This observation highlights the inherent advantage of rate coding, whose spikes are more informative than burst coding. Therefore, it can reserve more information with fewer spike counts.

\section{Conclusion}
\label{sec:conclusion}
{In this paper, we proposed a hybrid neural coding and learning framework to solve pattern recognition tasks with SNNs. In particular, we demonstrated that by holistically designing the SNN to encompass hybrid neural coding schemes, we could leverage the benefits of different neural coding schemes to achieve optimal performance according to task requirements and environmental conditions. To overcome the challenges associated with training hybrid coding SNNs, we proposed a layer-wise learning method that was both highly effective and efficient in producing the desired neural representations over existing SNN learning methods. Our extensive experiments demonstrated that state-of-the-art image classification accuracies and sound localization precision could be achieved with significantly improved inference speed, energy efficiency, and noise robustness compared to other SNNs constructed using a homogeneous neural coding scheme.}

{While our study primarily examined image classification and sound localization tasks, we emphasized that the design principles and training methods proposed in this work could be extended to other pattern recognition tasks and diverse environmental conditions. Notably, the successful implementation of our hybrid coding approach necessitated a comprehensive understanding of the strengths and limitations of various neural coding schemes as well as task-specific requirements. In future research, we aimed to develop a neural coding search method that could automatically identify the optimal hybrid neural coding design for each individual spiking neuron according to task demands and environmental conditions, thereby reducing the requisite level of expertise.}

\bibliographystyle{IEEEtran}
\bibliography{reference}
\vspace{-10mm}

\begin{IEEEbiography}[{\includegraphics[width=1in,height=1.25in]{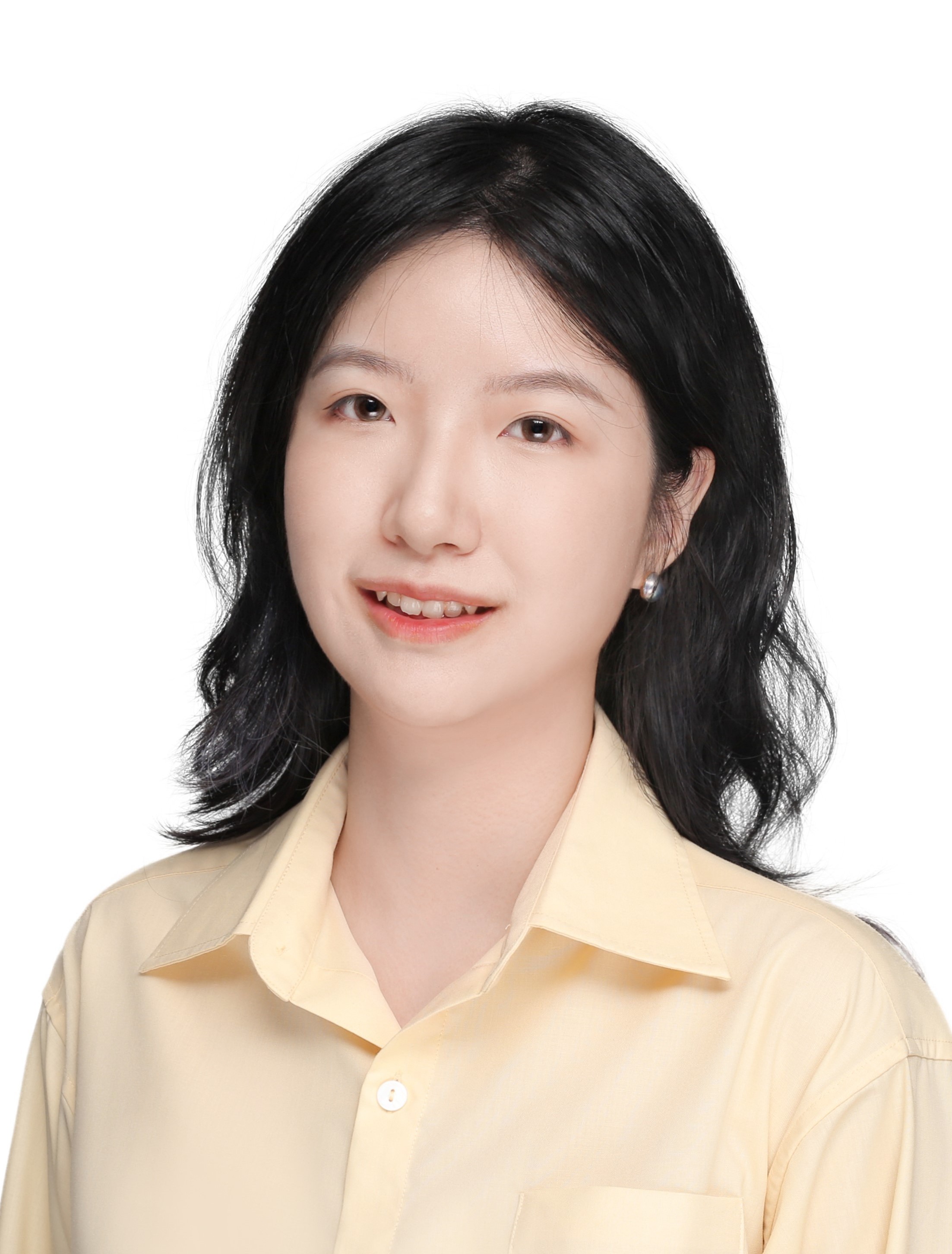}}]{Xinyi Chen} received the B.E. and M.Sc degree in Control Science and Engineering from Zhejiang University, China in 2019 and 2022, respectively. She is currently a Ph.D. student at the Department of Computing, the Hong Kong Polytechnic University. Her research interests include Brain-inspired Machine Intelligence, Spiking Neural Networks, and Neuromorphic Computing.
\end{IEEEbiography}
\vspace{-12mm}
\begin{IEEEbiography}[{\includegraphics[width=1in,height=1.25in]{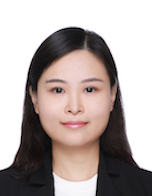}}]{Qu Yang} 
received a B.Eng. and an M.Sc. degree in electrical engineering from the National University of Singapore, Singapore in 2016 and 2019, respectively. She is currently a Ph.D. candidate in electrical engineering at the National University of Singapore. Her research focuses on neuromorphic computing and speech processing.
\end{IEEEbiography}
\vspace{-12mm}
\begin{IEEEbiography}[{\includegraphics[width=1in,height=1.25in]{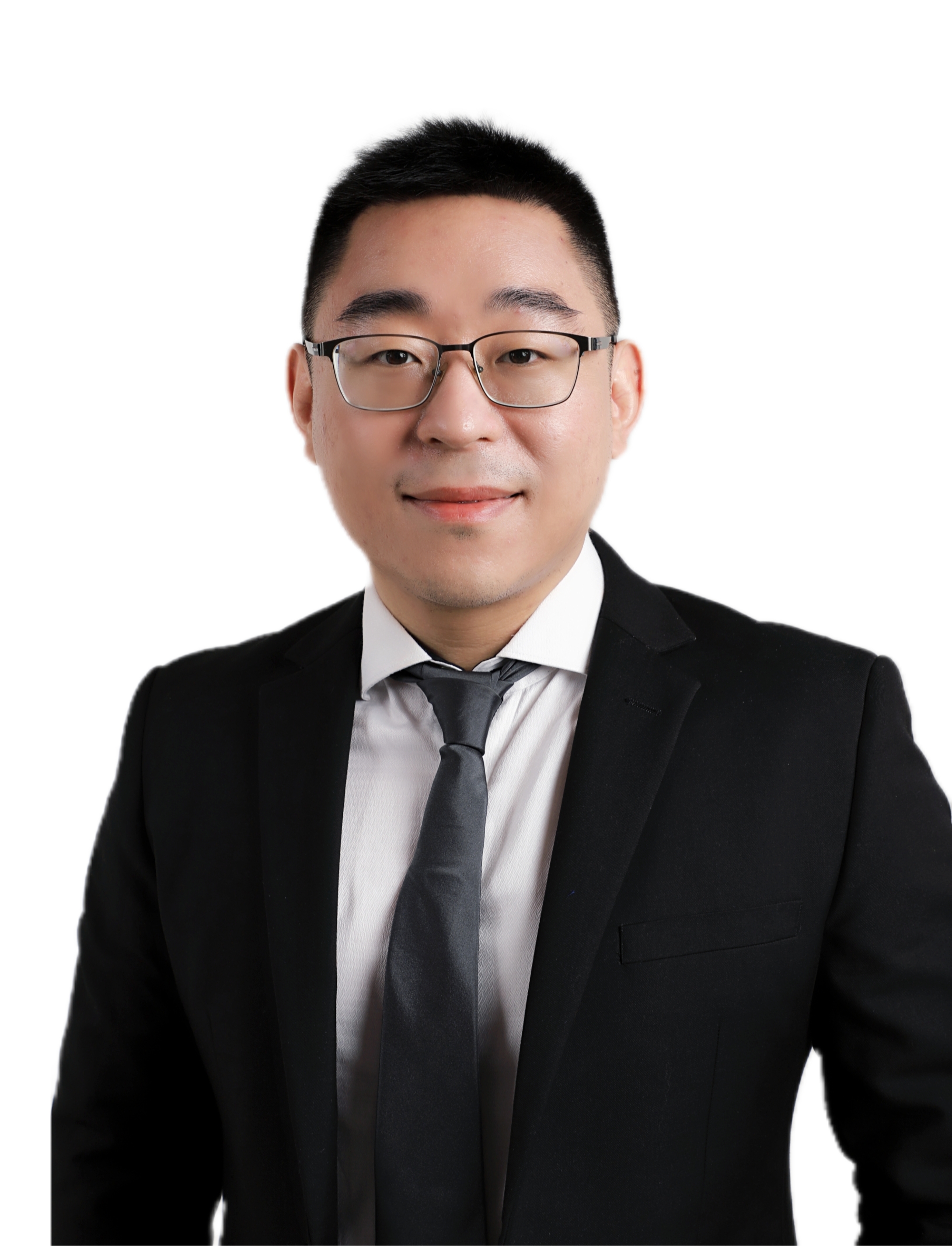}}]{Jibin Wu} received the B.E. and Ph.D degree in Electrical Engineering from National University of Singapore, Singapore in 2016 and 2020, respectively. Dr. Wu is currently an Assistant Professor in the Department of Computing, the Hong Kong Polytechnic University. His research interests broadly include brain-inspired artificial intelligence, neuromorphic computing, computational audition, speech processing, and machine learning. 
Dr. Wu has published over 30 papers in prestigious conferences and journals in artificial intelligence and speech processing, including NeurIPS, AAAI, TPAMI, TNNLS, TASLP, Neurocomputing, and IEEE JSTSP. He is currently serving as the Associate Editors for IEEE Transactions on Neural Networks and Learning Systems and IEEE Transactions on Cognitive and Developmental Systems. He also serves as the Editor for the Natural Language Processing Journal.
\end{IEEEbiography}
\vspace{-12mm}

\begin{IEEEbiography}[{\includegraphics[width=1in,height=1.25in]{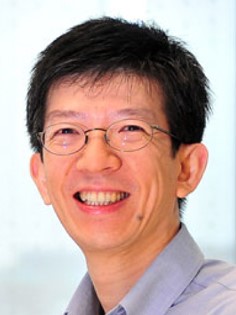}}]{Haizhou Li} 
(Fellow, IEEE) received the B.Sc., M.Sc., and Ph.D. degrees in electrical and electronic engineering from the South China University of Technology, Guangzhou, China, in 1984, 1987, and 1990 respectively. He is currently a Presidential Chair Professor and the Executive Dean of the School of Data Science, The Chinese University of Hong Kong, Shenzhen, China. He is also an Adjunct Professor with the Department of Electrical and Computer Engineering, National University of Singapore, Singapore. Prior to that, he taught with The University of Hong Kong, Hong Kong, during 1988–1990, and South China University of Technology, during 1990–1994. He was a Visiting Professor with CRIN, France, during 1994–1995, the Research Manager with the AppleISS Research Centre during 1996–1998, Research Director with Lernout \& Hauspie Asia Pacific during 1999–2001, Vice President with InfoTalk Corp. Ltd., during 2001–2003, and Principal Scientist and Department Head of human language technology with the Institute for Infocomm Research, Singapore, during 2003–2016. His research interests include automatic speech recognition, speaker and language recognition, and natural language processing. Dr. Li was the Editor-in-Chief of IEEE/ACM TRANSACTIONS ON AUDIO, SPEECH AND LANGUAGE PROCESSING during 2015–2018, has been a Member of the Editorial Board of Computer Speech and Language since 2012, an elected Member of IEEE Speech and Language Processing Technical Committee during 2013–2015, the President of the International Speech Communication Association during 2015–2017, President of Asia Pacific Signal and Information Processing Association during 2015–2016, and President of Asian Federation of Natural Language Processing during 2017–2018. He was the General Chair of ACL 2012, INTERSPEECH 2014, ASRU 2019, and ICASSP 2022. Dr. Li is a Fellow of the ISCA, and a Fellow of the Academy of Engineering Singapore. He was the recipient of the National Infocomm Award 2002, and President’s Technology Award 2013 in Singapore. He was named one of the two Nokia Visiting Professors in 2009 by the Nokia Foundation, and U Bremen Excellence Chair Professor in 2019.
\end{IEEEbiography}
\vspace{-12mm}
\begin{IEEEbiography}[{\includegraphics[width=1in,height=1.25in]{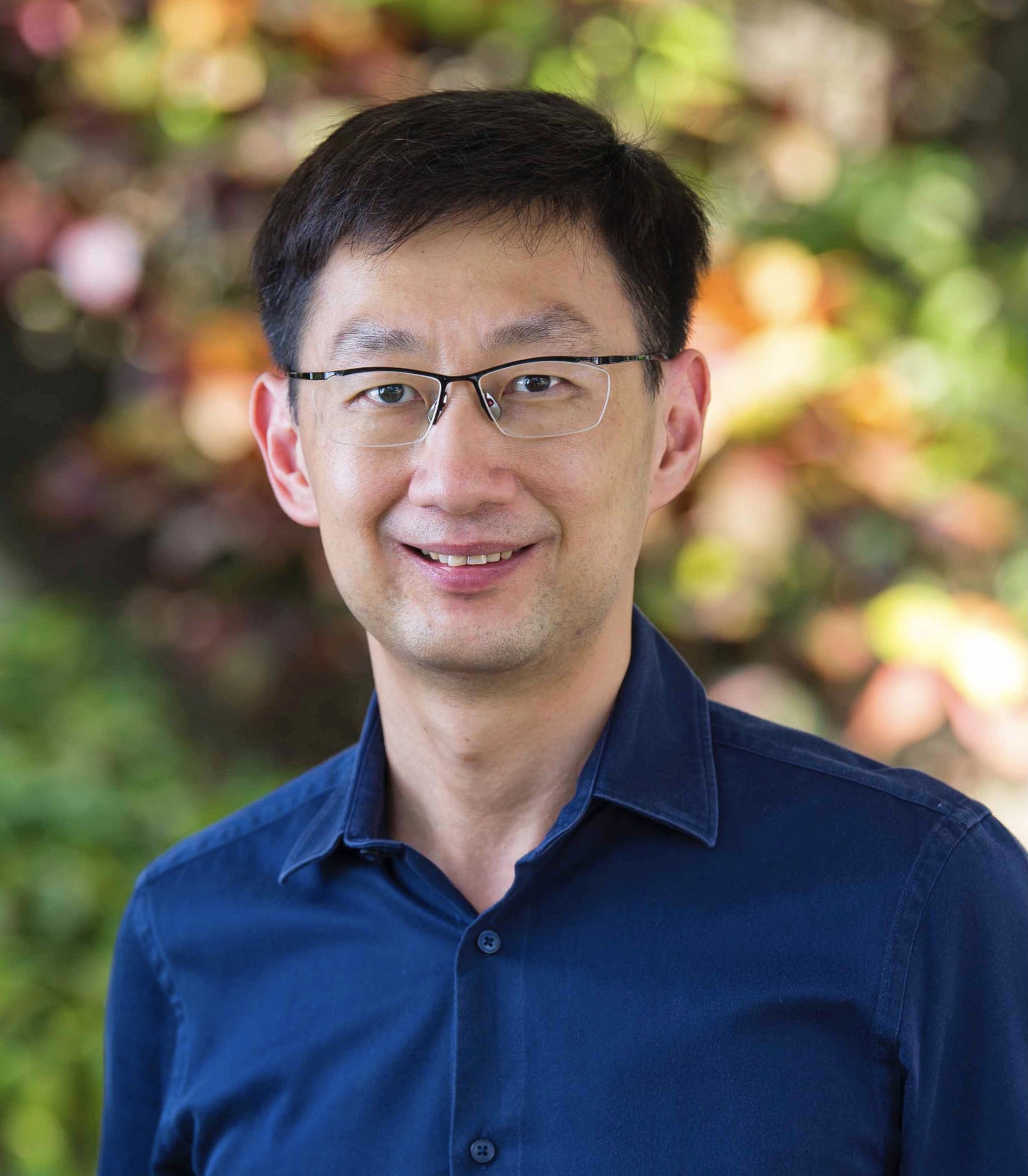}}]{Kay Chen Tan} (Fellow, IEEE) received the B.Eng. degree (First Class Hons.) and the Ph.D. degree from the University of Glasgow, U.K., in 1994 and 1997, respectively. He is currently a Chair Professor (Computational Intelligence) of the Department of Computing, the Hong Kong Polytechnic University. He has published over 300 refereed articles and seven books. 
Prof. Tan is currently the Vice-President (Publications) of IEEE Computational Intelligence Society, USA. He has served as the Editor-in-Chief of the IEEE Computational Intelligence Magazine from 2010 to 2013 and the IEEE TRANSACTIONS ON EVOLUTIONARY COMPUTATION from 2015 to 2020, and currently serves as the Editorial Board Member for more than ten journals. He is the Chief Co-Editor of Springer Book Series on Machine Learning: Foundations,
Methodologies, and Applications.
\end{IEEEbiography}

\onecolumn
\pagestyle{plain}
\clearpage
\newpage
\setcounter{page}{1}

\Large
\begin{center}
    {\bf \LARGE Supplementary Materials}\\
    \vspace{0.2cm}
    ``{\large A Hybrid Neural Coding Approach for Pattern Recognition with Spiking Neural Networks}"\\
	\vspace{0.2cm}
    {\normalsize Xinyi Chen{*}, Qu Yang{*}, Jibin~Wu,
	Haizhou~Li,~\IEEEmembership{Fellow,~IEEE},	  Kay~Chen~Tan,~\IEEEmembership{Fellow,~IEEE}}\\
\end{center}
\rule[-0.5pt]{18.1cm}{0.06em}

\parskip=2pt

\large

\parskip=2pt

\noindent
\subsection{\large Implementation Details}
Our experiments are conducted on the CIFAR-10, CIFAR-100, Tiny-ImageNet, and SLoClas datasets. For the image classification datasets, we employ both VGG-16 \cite{simonyan2014very} and ResNet-20 \cite{he2016deep} network structures. For the sound localization dataset, we use network architecture as defined in Table \ref{tab:config}. 

\subsubsection{\large Datasets}
\noindent
\textbf{CIFAR-10} \cite{krizhevsky2009learning} This dataset contains 60,000 colored images from 10 classes. Each of the images has a size of 32 $\times$ 32 $\times$ 3. All the images are split into 50,000 and 10,000 for training and testing, respectively.

\noindent
\textbf{CIFAR-100} \cite{krizhevsky2009learning} This dataset contains 60,000 colored images from 100 classes. Each of the images has a size of 32 $\times$ 32 $\times$ 3. All the images are split into 50,000 and 10,000 for training and testing, respectively.

\noindent
\textbf{Tiny-ImageNet} \cite{wu2017tiny} This dataset contains 110,000 colored images from 200 classes. Each of the images has a size of 64 $\times$ 64 $\times$ 3. All the images are split into 100,000 and 10,000 for training and testing, respectively.

\noindent
\textbf{SLoClas} \cite{qian2021sloclas} This dataset contains 23.27 hours of audio recorded by a 4-channel microphone array at a distance of 1.5 meters. During data collection, 10 classes of sound events are played continuously, spanning 0$-$360 degrees with an interval of 5 degrees. 

\subsubsection{\large Data Preprocessing}
\noindent
For all image classification datasets, the data preprocessing techniques, including resize, random crop, random horizontal flip, and data normalization, follow the techniques described in \cite{QCFS}. More details can be found in our released code.

\noindent

In the task of sound localization, preprocessing of the raw audio is a necessary step before feeding it into the networks. The input to neural networks can either be at the signal level or the feature level. The majority of methods listed in Table \ref{tab:SL_results} \cite{he18b_interspeech,adavanne2018sound, EINV2, yang2022srpdnn, wang2023fnssl} adopt the short-time Fourier transform (STFT) \cite{stft}, a commonly used feature extraction method in the frequency domain, to process the raw audio signal into STFT coefficient. Subsequently, the real and imaginary parts of the STFT coefficients (or alternatively, the magnitude and phase) are directly used as the network input. In this work, we adopt the spike phase coding to extract the inter-channel phase difference (IPD) features from the STFT-encoded data, as demonstrated in Fig. \ref{fig: prepocess}, with the configuration detailed in Table \ref{tab:config}. It's important to note that, for a fair comparison, we employ the same STFT configuration across all methods utilized STFT. Specifically, the sampling rate is 16 kHz, the window length and frame stride are both $170$ ms, the FFT length is 1024, and the Hamming window is used.

\begin{table}[h]
\normalsize
\centering
\caption{Experimental configuration of the sound localization task. }
\label{tab:config}
\resizebox{0.8\textwidth}{!}{%
\begin{tabular}{l  c }
\hline
\hline
 \textbf{Attributes} & \textbf{Setup} \\
 \hline
 \textbf{1. Data prepossessing:} \\
 Sampling rate & 16 kHz \\
 Frame length & 170 ms \\
 Frame stride & 170 ms\\
 FFT points $n$ & 1024 \\
 Number of Microphones & 4 \\
 \hline
 \textbf{2. MTPC front-end:} \\
 Frequency channels $N$ & 40 \\
 Coincidence detector $K_{\tau}$ & 51 \\
 Microphone pairs $K$ & 6 \\
 CQT frequency range & [280, 8000] \\
 \hline
\textbf{3. SNN architecture:}  \\
 Hidden Layers  & 24c3s2p1-48c3s2p1-96c3s2p1-fc1024* \\
 Output Layer & fc360\\ 
\hline
\hline
\multicolumn{2}{l}{* c - convolution kernel size, s - stride, p - padding, fc - fully-connected.} \\
\multicolumn{2}{l}{* For example, 24c3s2p1 represents a layer with 24 channel, 3*3 convolution kernel, stride 2, and padding 1.}
\end{tabular}}
\end{table}

\subsubsection{\large Computing Infrastructure}
All experiments and data analysis are performed on a computer equipped with Ubuntu 20.04.4 LTS, Intel Xeon E5-2640 CPU, GeForce GTX 1080Ti GPUs (11G Memory), and PyTorch 1.11.0. 

\subsection{\large Ablation Studies}
Here, we perform several ablation studies to empirically explore the impact of different hyperparameters on the performance of the proposed hybrid coding SNNs. {Our findings indicate that the set of hyperparameters that yields optimal performance can be effectively transferred across different tasks. Drawing on these observations, we can offer a suggested hyperparameter setting strategy for effortless tuning of these hyperparameters when implemented in diverse tasks}. All experiments in this section are conducted on the CIFAR-10 dataset with VGG-16 network architecture, {the CIFAR-100 dataset with VGG-16 architecture, and the SLoClas dataset with the architecture described in Table \ref{tab:config}.}
\subsubsection{\large Scaling Factor in Local Tandem Learning (LTL)}
{To study the influence of the scaling factor $\boldsymbol{r}^l$ in Eq. (\ref{eq: LTL_loss}) of the manuscript, we conduct experiments by progressively increasing the value of $\boldsymbol{r}^l$ while keeping all other hyperparameters fixed at a well-established setting. Specifically, the maximum spike count in burst coding is set to 5, while direct coding is employed for the output layer. The results of these experiments are presented in Table \ref{tab: scaling_factor}. Our findings reveal that the performance of the model remains largely unaffected when the scaling factor $\boldsymbol{r}^l$ is less than or equal to 1. However, beyond this threshold, there is a noticeable decrease in accuracy for both the CIFAR-10 and CIFAR-100 tasks. This decline can be attributed to the excessive scaling of the SNN's firing rate, causing a significant mismatch with the teaching signals from the ANN. \textbf{Based on these observations, it is recommended to set the value of $\boldsymbol{r}^l$ to 1}.}

\begin{table}[h]
  \caption{Study of accuracy (\%) on the scaling factor $\boldsymbol{r}^l$ across different tasks.}
  \label{tab: scaling_factor}
  \centering
  \resizebox{0.8\textwidth}{!}{%
  \begin{tabular}{lcccccccc}
\hline
    \textbf{Scaling factor ($\boldsymbol{r}^l$)}  & 0.2  &0.4 & 0.6 &0.8 &1 &2 &3 &4 \\
    \hline
     \textbf{CIFAR-10}     & 95.59  & 95.71 & 95.68 & 95.58 & 95.53 & 94.74 & 93.33 & 91.73 \\
    \hline
     \textbf{CIFAR-100}     & 76.44  & 76.75 & 76.79 & 76.60 & 76.37 & 73.66 & 69.70 & 64.22 \\
    \hline
    \textbf{SLoClas} & 95.95 & 95.95 & 95.94 & 95.95 & 95.94 & 96.03 & 95.95 & 96.05 \\
    \hline
  \end{tabular}}
\end{table}

\subsubsection{\large Maximum Spike Count in Burst Coding}
{Next, we conduct experiments to investigate the influence of the maximum spike count, denoted as $\Gamma$, in burst coding as described in Eq. (\ref{eq: burst}) of the manuscript. Same as the prior study on the scaling factor $\boldsymbol{r}^l$, we progressively increase the value of $\Gamma$ while keeping all other hyperparameters fixed at a well-established setting. Specifically, the scaling factor $\boldsymbol{r}^l$ is set to 1, and the direct coding is employed for the output layer. The results, presented in Table \ref{tab:gamma}, suggest that this particular hyperparameter has a \textbf{negligible effect} on the model's performance when the spike count is greater than or equal to 2. This phenomenon can be attributed to the saturation of the required information transmission rate. Although the addition of more spikes may lead to marginal improvements, the conveyed information is already sufficiently captured by the first few spikes. \textbf{Based on these observations, it is recommended to set the value of $\Gamma$ to 5}. This choice not only aligns with the biological finding that neurons seldom emit spike bursts exceeding five spikes \cite{buzsaki2012neurons}, but also adheres to our experimental findings that an increase in spike count beyond 5 yields negligible improvements in task performance.}

\begin{table}[h]
  \caption{Study on the accuracy (\%) of maximum spike count $\Gamma$ in burst coding across different tasks.}
  \label{tab:gamma}
  \centering
  \resizebox{1.0\textwidth}{!}{%
  \begin{tabular}{lcccccccccc}
\hline
    \textbf{Maximum spike count ($\Gamma$)} &1 & 2 & 3 & 4 & 5 & 6 & 7 & 8 & 9 & 10 \\
    \hline
    \textbf{CIFAR-10}       &95.26 & 95.50 & 95.55 & 95.59 & 95.53 & 95.57 & 95.56 & 95.58 & 95.46 & 95.43 \\
    \hline
    \textbf{CIFAR-100}       &74.65 & 76.13 & 76.35 & 76.52 & 76.37 & 76.28 & 76.26 & 76.27 & 76.14 & 76.14 \\
    \hline
    \textbf{SLoClas} & 95.22 & 95.65 & 95.86 & 95.96 & 95.94 & 95.95 & 95.90 & 95.95 & 95.94 & 95.97\\
    \hline
  \end{tabular}}
\end{table}

\subsubsection{\large Loss Weights in Time-to-First-Spike (TTFS) Learning}
\begin{figure}[h]
    \centering
    \subfloat[CIFAR-10]{
	\includegraphics[scale=0.49,trim=10 0 0 0,clip]{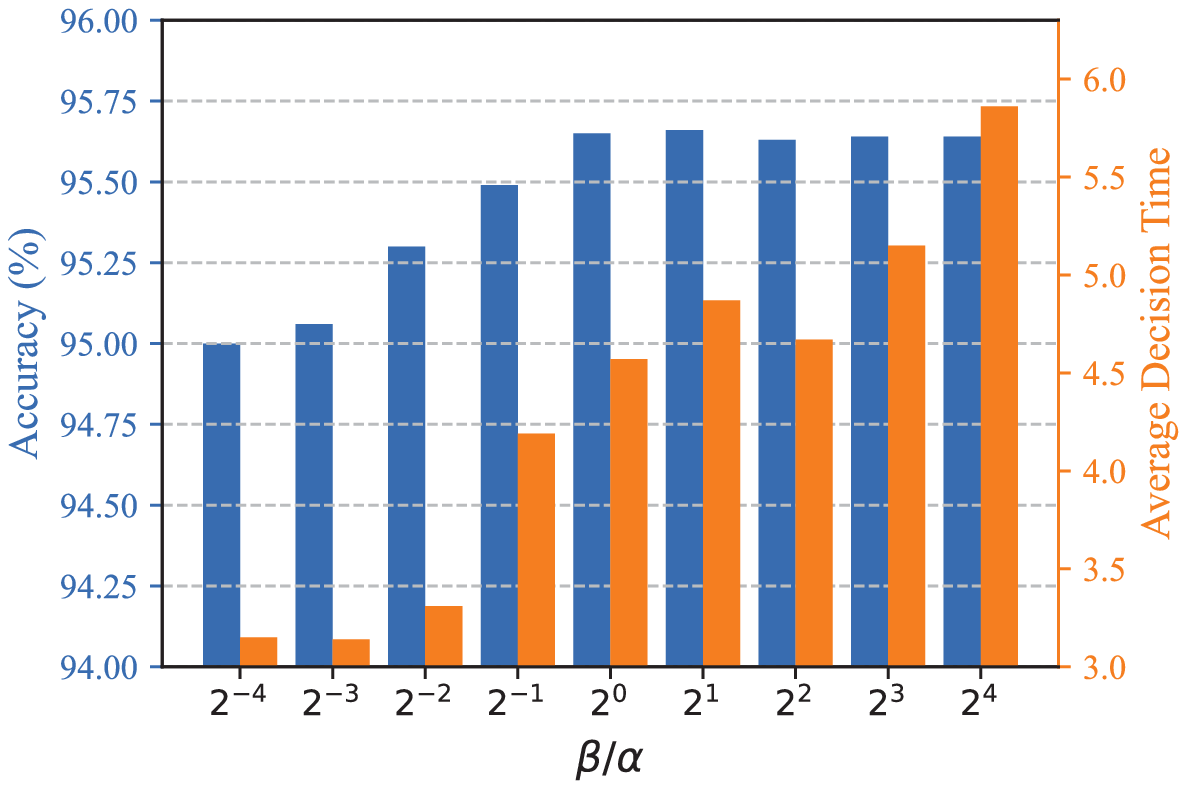}}
 \subfloat[CIFAR-100]{
	\includegraphics[scale=0.49,trim=10 0 0 0,clip]{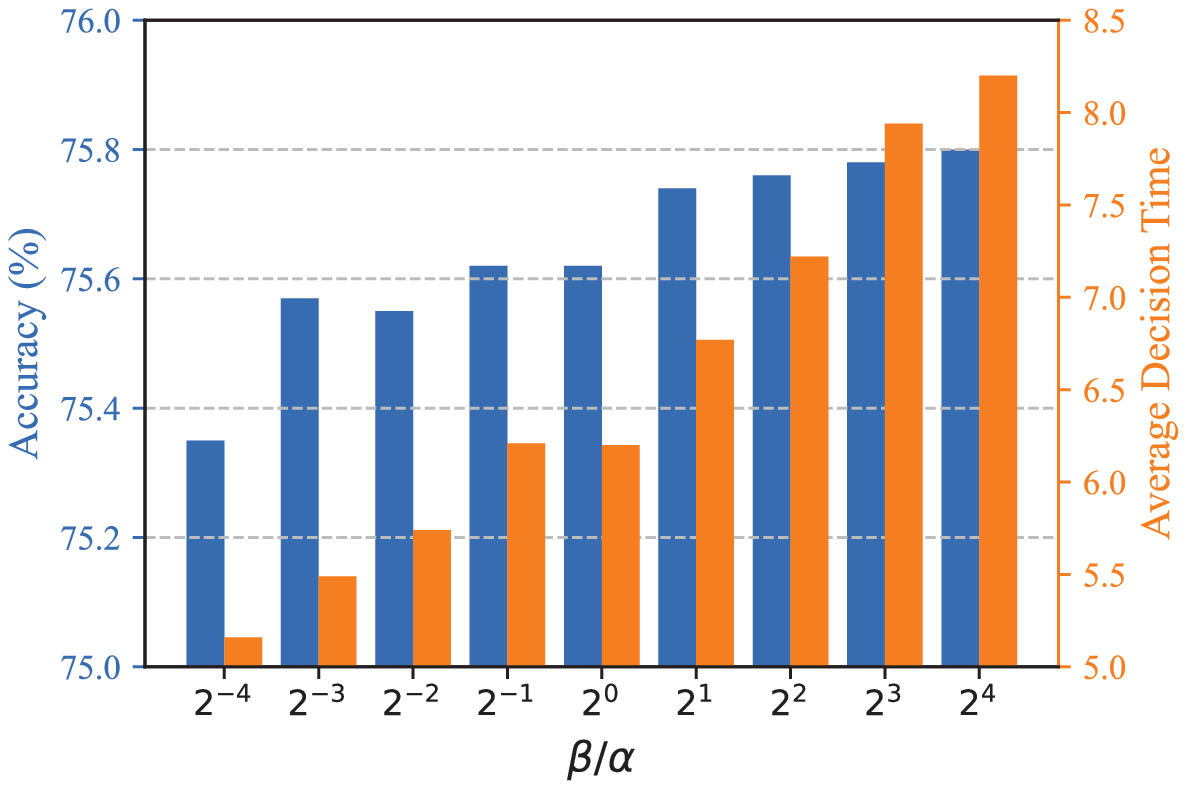}}
 \subfloat[SLoClas]{
	\includegraphics[scale=0.49,trim=10 0 0 0,clip]{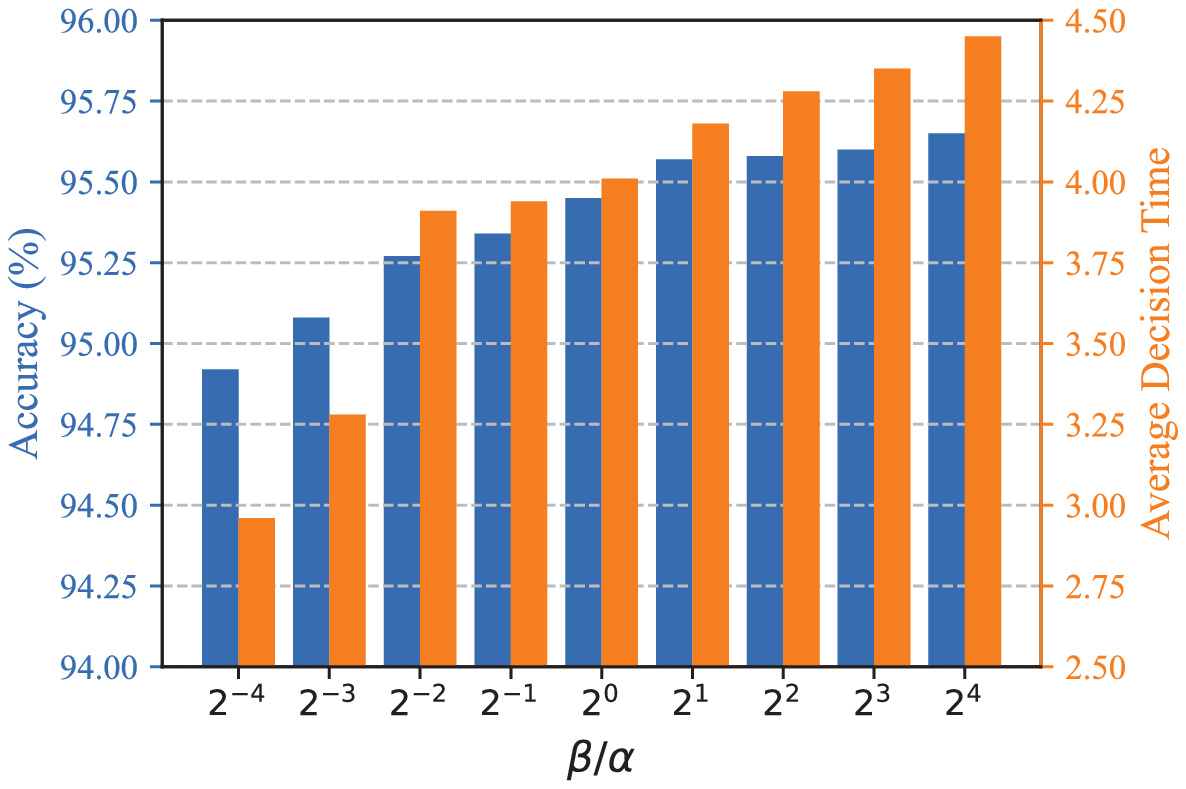}}
    \caption{Study on the impact of the ratio between two loss functions $\beta/\alpha$ across different tasks.}
    \label{fig:abl_ttfs}
\end{figure}

{As described in Eqs. (\ref{loss1}) and (\ref{loss2}) of the manuscript, the training of the output layer incorporates two losses, denoted as $\mathcal{L}^L_1$ and $\mathcal{L}^L_2$. These losses are introduced to strike a balance between classification accuracy and inference latency. To achieve this balance, hyperparameters $\alpha$ and $\beta$ are employed to trade-off between the two loss functions, resulting in a combined loss function of $\mathcal{L}^L=\alpha \mathcal{L}^L_1+\beta \mathcal{L}^L_2$. In order to evaluate the impact of the ratio between $\alpha$ and $\beta$ on task performance, we hold $\alpha$ at a constant value of 2 while varying the value of $\beta$. The results, as demonstrated in Fig. \ref{fig:abl_ttfs}, consistently reveal trade-off patterns between accuracy and inference latency across all three tasks. Specifically, as the $\beta/\alpha$ ratio increases, the average decision time monotonically increases, leading to a noticeable enhancement in classification accuracy. However, once the $\beta/\alpha$ ratio exceeds a certain threshold (larger than 2), the delay in decision time no longer yields a significant improvement in accuracy. This observation suggests the gradual saturation of network accuracy. \textbf{Based on these findings, it is recommended to set the $\beta/\alpha$ ratio based on the specific requirements of individual tasks. For instance, if high accuracy is of paramount importance, a larger $\beta/\alpha$ should be selected. On the other hand, if minimizing inference latency is the primary objective, a smaller $\beta/\alpha$ would be more appropriate. In our study, we employ the metric $\Delta \text{Acc}. \times t_{inf}$ as a means to attain performance that trade-off between accuracy and inference latency.}}

\subsection{\large Experimental Settings}
\subsubsection{\large Hyperparameters}

The hyperparameters implemented for different datasets and network structures are summarized in Table \ref{tab:hyperpara}. For the hidden layers, the firing threshold $V_{th}^l$ is fixed to be 1 for all bursting IF neurons. $\boldsymbol{r}^l$ is set to 1, and $\Gamma$ is set to 5 according to our conclusion derived from ablation studies. For the output layer, the firing threshold $V_{th}^L$ used for TTFS coding is determined by the mean output membrane potential, varying with different tasks. The decay constants of all LIF neurons, namely $\tau_m$, $\tau_s$, are fixed to 2.0 and 0.5 following the setting in \cite{gutig2006tempotron}. $\alpha$ and $\beta$ are the coefficients employed to trade off the TTFS loss functions, whose values vary slightly with different datasets.

\begin{table}[h]
\normalsize
\centering
\caption{Hyperparameter Settings Used for the Reported Results of Hybrid Coding Models.}
\label{tab:hyperpara}
\resizebox{0.7\textwidth}{!}{%
\begin{tabular}{l c c c c c c c c c}
\hline
\hline
\textbf{Dataset}  & \textbf{Architecture} &\textbf{$V_{th}^l$} &\textbf{$\Gamma$} &\textbf{$\boldsymbol{r}^l$}  &\textbf{$V_{th}^L$}  &\textbf{$\tau_m$} &\textbf{$\tau_s$}  &\textbf{$\alpha$} &\textbf{$\beta$}  \\ 
\hline
CIFAR-10        &VGG-16 &1 &5 & 1 &3   &2 &0.5  &2    &2   \\
CIFAR-10        &ResNet-20 &1 &5  & 1 &3   &2 &0.5      &2    &1   \\
CIFAR-100       &VGG-16 &1 &5 & 1 &8.5    &2 &0.5    &2    &0.4   \\
CIFAR-100       &ResNet-20 &1 &5 & 1 &6.5    &2 &0.5      &2    &0.2   \\
Tiny-ImageNet   &VGG-16 &1 &5 & 1   &4.5     &2 &0.5     &2    &1   \\
Tiny-ImageNet   &ResNet-20 &1 &5 & 1  &4.5   &2 &0.5     &2    &2   \\
SLoClas & Our Model &1 &5 & 1 & 4.5 & 2 & 0.5 & 2 & 5 \\

\hline
\hline

\end{tabular}}
\end{table}

\subsubsection{\large Training Configurations}
We pre-train all teacher ANNs for 300 epochs using the SGD optimizer with a momentum of 0.9 and weight decay of $5\times10^{-4}$. The initial learning rates are 0.1 for CIFAR-10/CIFAR-100, 0.01 for Tiny-ImageNet, and 0.5 for SLoClas. We use the Cosine Annealing scheduler for learning rate adjustment. 

The initial membrane potential of all SNNs is set to 0. For hidden layers, we fine-tune the student SNNs with the Adam optimizer for 100 epochs for CIFAR-10/CIFAR-100/SLoClas and 50 epochs for Tiny-ImageNet. The initial learning rate is set to $1\times 10^{-4}$ and decays by 0.5 every 10 epochs for CIFAR-10/CIFAR-100/SLoClas and 0.2 every five epochs for Tiny-ImageNet. For the TTFS output layer, it is trained with the SGD optimizer for 50, 50, 100, and 100 epochs for CIFAR-10, CIFAR-100, Tiny-ImageNet, and SLoClas, respectively. The initial learning rate is set to $1\times 10^{-5}$ for all SNNs and decays its value by 0.5 at Epoch 15, 25, and 40 for CIFAR-10/CIFAR-100, at Epoch 30, 50, and 80 for Tiny-ImageNet, and at Epoch 20, 40, 60, 80 for SLoClas, respectively. 

\subsubsection{\large Implementation Details of Rate Coding and Direct Coding}
In Fig. \ref{fig:distri}, Fig. \ref{fig: 4in1}, and Section \ref{sec:ablation}, rate coding is employed as another candidate coding scheme for hidden layers. The implementation of rate coding in our framework is highly similar to burst coding, as the main difference between rate coding and burst coding neurons is that rate coding neurons can emit at most one spike at a time, whereas bursting neurons can generate up to $\Gamma$ spikes. Therefore, we could easily derive the model of rate coding neurons by simply setting $\Gamma$ in Eq. (\ref{eq: burst}) to 1. 

Furthermore, direct coding is also adopted in Fig. \ref{fig: 4in1} and Section \ref{sec:ablation} as a promising selection of output coding scheme in our framework. In this design, we adhere to the convention followed by other SNNs \cite{QCFS,wu2021tandem}. It decodes the network output by averaging the output of a linear classifier across all timesteps, which is defined below: 
\begin{equation}
\boldsymbol{y}^L=\frac{1}{T}\sum^T_{t=1}{\mathcal{W}^L\boldsymbol{s}^{L-1}(t)},    
\end{equation} 
where $\boldsymbol{s}^{L-1}$ signifies the output spikes of the last hidden layer $L-1$, which serves as the input to the output layer $L$. $\mathcal{W}^L$ is the weight matrix of the output linear classifier, while $T$ denotes the time window length. Unlike the meticulously designed loss functions used by TTFS output coding, the models that utilize direct coding incorporate the cross-entropy (CE) loss function during the training process.

\end{document}